\documentclass{article}

\usepackage[utf8]{inputenc}
\usepackage[T1]{fontenc}
\usepackage{graphicx}
\usepackage{amsmath, amsfonts, amssymb}
\usepackage{xcolor}
\usepackage{hyperref} % For clickable email links
\usepackage{geometry}
\usepackage{array}
\usepackage{makecell}
\usepackage{multirow}
\usepackage{rotating}
\usepackage{caption}
\usepackage{subcaption}
\usepackage[para,online,flushleft]{threeparttable}
\usepackage{float}
\floatstyle{ruled}

\usepackage{units}
\usepackage{diagbox}
\usepackage[normalem]{ulem}
\usepackage{lineno}
\usepackage{etoolbox}
\usepackage{blindtext} % For placeholder text
\usepackage{comment}

\usepackage{natbib}
\usepackage{algorithm}
\usepackage{algpseudocode}

\usepackage{tikz}
\usetikzlibrary{shapes, arrows, positioning}

\providecommand{\keywords}[1]
{
  \small	
  \textbf{\textit{Keywords---}} #1
}
\begin{document}

% --- MANUAL TITLE PAGE ---
% The 'titlepage' environment ensures this is a separate page with no page number.
\begin{titlepage}
    \centering % Center all content on the page

    % --- Title ---
    {\huge \bfseries MoCap-Impute: A Comprehensive Benchmark and Comparative Analysis of Imputation Methods for IMU-based Motion Capture Data\par}
    
    \vspace{2.5cm} % Add significant vertical space after the title

    % --- Author Block ---
    % Minipage contains the author list, allowing it to line-break if needed
    \begin{minipage}{\textwidth}
        \centering
        \large % Set author font size
        Mahmoud Bekhit\textsuperscript{1,2,*},
        Ahmad Salah\textsuperscript{3,4,*,†},
        Ahmed Salim Alrawahi\textsuperscript{3,5,*},
        Tarek Attia\textsuperscript{6},
        Ahmed Ali\textsuperscript{7,8,†},
        Esraa Eldesokey\textsuperscript{7,9},
        Ahmed Fathalla\textsuperscript{10}
    \end{minipage}\par
    
    \vspace{1.5cm} % Add vertical space after the author list

    % --- Affiliation Block ---
    \normalsize
    \begin{tabular}{l}
        \textsuperscript{1}Peter Faber Business School, Australian Catholic University (ACU), Sydney, Australia\\
        \textsuperscript{2}
        University of Technology Sydney (UTS), Sydney, NSW 2007, Australia \\
        \textsuperscript{3}Department of Information Technology, College of Computing and Information Sciences,\\ University of Technology and Applied Sciences, Ibri 511, Oman \\
        \textsuperscript{4}Faculty of Computers and Informatics, Zagazig University, Zagazig, Egypt \\
        \textsuperscript{5}AI Applications Research Chair, University of Nizwa, Nizwa 616, Oman\\
        \textsuperscript{6}Department of Data Science, Wakeb, Riyadh, KSA\\
        \textsuperscript{7}Department of Computer Science, College of Computer Engineering and Sciences, \\Prince Sattam Bin Abdulaziz University, Alkharj, Saudi Arabia.\\
        \textsuperscript{8}Higher Future Institute for Specialized Technological Studies, Cairo, Egypt.\\
        \textsuperscript{9}Department of Computer Science, Faculty of Computers \& Informatics, Suez Canal University, Ismailia, Egypt\\
        \textsuperscript{10}Department of Mathematics, Faculty of Science, Suez Canal University, Ismailia, Egypt\\[2ex] % Adds a bit of space before the footnotes
        \textsuperscript{*}These authors contributed equally to this work. \\
        \textsuperscript{†}Corresponding authors: \href{mailto:ahmad.salah@utas.edu.om}{ahmad.salah@utas.edu.om}, \href{mailto:a.ali@psau.edu.sa}{a.abdalrahman@psau.edu.sa}
    \end{tabular}

    \vfill % Fills the remaining vertical space, pushing the date to the bottom
\end{titlepage}

% --- MAIN DOCUMENT BEGINS HERE ---

\begin{abstract}

Motion capture (MoCap) data from wearable Inertial Measurement Units (IMUs) is vital for applications in sports science and healthcare, yet its utility is often compromised by missing data. Despite numerous imputation techniques, a systematic performance evaluation for IMU-derived MoCap time-series data is lacking. We address this gap by conducting a comprehensive comparative analysis of statistical, machine learning, and deep learning imputation methods. Our evaluation considers three distinct contexts: univariate time-series, multivariate across subjects, and multivariate across kinematic angles.

To facilitate this benchmark, we introduce the first publicly available MoCap dataset designed specifically for imputation, featuring data from 53 karate practitioners. We simulate three controlled missingness mechanisms: missing completely at random (MCAR), block missingness, and a novel value-dependent pattern at signal transition points.

Our experiments, conducted on 39 kinematic variables across all subjects, reveal that multivariate imputation frameworks consistently outperform univariate approaches, particularly for complex missingness. For instance, multivariate methods achieve up to a 50\% mean absolute error reduction (MAE from 10.8 to 5.8) compared to univariate techniques for transition point missingness. Advanced models like Generative Adversarial Imputation Networks (GAIN) and Iterative Imputers demonstrate the highest accuracy in these challenging scenarios. This work provides a critical baseline for future research and offers practical recommendations for improving the integrity and robustness of Mo-Cap data analysis.
\end{abstract}

\keywords{
Benchmark Dataset, Data Imputation, Deep Learning, Inertial Measurement Units (IMU), Machine Learning, Missing Data, Motion Capture, Performance Evaluation, Time-Series Analysis.}

% Dr. Gamal, Dr (Salah*, rawahi), Tarek, Dr. Ali*, Dr. Esraa, Dr. Fathalla

\section{Introduction}
MoCap has become a vital aspect of studying various human activities as in sports science \cite{van2018accuracy,fathalla2023real} or human disorders as in Neurosciences \cite{jalata2021movement}. This includes capturing gestures using Inertial Measurement Units (IMUs) sensors to study various human movements. IMUs wearable sensors are used to capture sport skills performed by players for further study and analysis such as human performance assessment \cite{arlotti2022benefits,adel2022survey}. 

%%%%%%%%%%%%%%%%%%%% Task 3 %%%%%%%%%%
IMUs sensors have recently acquired popularity due to their widespread use in wearable devices that aid in the detection of body parts motion and orientations. They have the ability to offer acceptable data rates and provide digital outputs, together with their reasonable cost and extended lifetime. Several applications have been encountered the use of IMUs including capture and monitor the movement of athletes in order to assess their talents and document their professional motions \cite{guignard2021validity}. In addition, they have been used to healthcare difficulties, such as neurological illnesses, where they are employed in daily activities and environments for remote diagnosis and rehabilitation direction \cite{wang2020two}. Additional uses include professional motion capture studios and intensive 3D animation and design resources \cite{karuzaki2021realistic}. During the collection of digital data, however, some data may be lost owing to the battery life of sensors or inadequate network connectivity. Likewise, the presence of metallic items in the surrounding of the IMUs might impact the accuracy of collected data.

To our knowledge, despite the benefits of IMUs sensors to a wide range of applications, its application to data imputation obtained from IMU sensors remains underexplored. The IMU sensors' readings are impacted negatively by several factors including drifting, placement of the sensors, and magnetic field interference \cite{yahyamotion}. These issues result in missing data which may lead to corrupted data, inaccurate outcomes or biased results. Data imputation is therefore required early in pre-processing stage to treat missing data. 
This study investigates the efficacy of typical data imputation techniques when applied to MoCap data acquired from IMUs wearable sensors. So far, the existing literature lacks the comprehensive performance of data imputation mainly established to this sensory data. As a consequence, this study provides the first  MoCap-based dataset for benchmarking the data imputation various methods and then report on the performance of data imputation techniques. The issue of missing data will be evaluated for MoCap data. The utilized models in this study belong to three major categories, namely, machine learning (ML), deep learning (DL), and statistical methods. The key  contributions of this study are listed as follows:

\begin{enumerate}
\item To our knowledge, this is the first work to explore the missing data imputation for MoCap data gathered from IMUs wearable sensors.

\item To our knowledge, the first MoCap dataset that was proposed to be utilized for the sake of imputing the missing values.

\item Several well-known missing data imputation methods were tested and recommended the most suitable techniques for MoCap missing data collected from IMUs.
\end{enumerate}

The rest of this paper is organized as follows. In Section \ref{sec:related-work}, the related work of the missing data imputation methods is addressed. The proposed dataset along with the proposed methodology and data evaluation metrics are presented in Section  \ref{sec:methodology}. Section \ref{sec:results} lists and discusses the study results. Finally, the paper is concluded in Section \ref{sec:conclusions}.

%latest draft by Ahmed Alrawahi
\section{Background and Related Work\label{sec:related-work}}

\subsection{IMUs}
IMUs sensors have gained popularity over the recent years in numerous applications, including manufacturing, robotics, healthcare, and evaluations \cite{clemente2021validity}. IMUs are portable electronic equipment used to track and measure the angular velocity and the body motion. There exist different forms of data collection sensors such as accelerometers, gyroscopes, and magnetometer. Accelerometers and gyroscopes measure the inertial acceleration and rotational angle. However, the magnetometer measures the bearing magnetic direction that enhances the acquired readings and is considered as an advanced sort of sensors. IMUs are considered one of the most straightforward and rapid means for capturing the motion of the body parts due to the omission of cable-extension during data acquisition \cite{ahmad2013reviews}. Since IMUs have been widely applied to determine motion in terms of acceleration, angular velocity, and orientation \cite{zhu2004real}, the accelerometer measures total acceleration $\mathbf{a}_m$ as shown in Eq. \ref{eq:1}.

\begin{equation} \label{eq:1}
\mathbf{a}_m = \mathbf{a}_b + \mathbf{g} + \boldsymbol{\eta}_a
\end{equation}
where $\mathbf{a}_b$ represents the body acceleration due to external forces, $\mathbf{g}$ represents the gravitational acceleration vector with magnitude 9.81 m/s², and $\boldsymbol{\eta}_a$ represents accelerometer noise.

The angular velocity $\boldsymbol{\omega}_m$ is measured in degrees per second using the gyroscope sensor:

\begin{equation} \label{eq:2}
\boldsymbol{\omega}_m = \boldsymbol{\omega}_t + \mathbf{b}_g + \boldsymbol{\eta}_g
\end{equation}
where $\boldsymbol{\omega}_t$ is the true angular velocity, $\mathbf{b}_g$ is the gyroscope bias, and $\boldsymbol{\eta}_g$ is gyroscope noise.

The relationship between orientation $\boldsymbol{\theta_{orintation}}(t)$ and true angular velocity is $\boldsymbol{\omega}_t(t) = d\boldsymbol{\theta_{orintation}}(t)/dt$. In discrete time, orientation is estimated by integrating measured angular velocity using Euler integration:

\begin{equation}
\boldsymbol{\theta_{orintation}}(t) \approx \boldsymbol{\theta_{orintation}}(t - \Delta t) + \boldsymbol{\omega}_m(t - \Delta t) \cdot \Delta t
\label{eq:angle_update}
\end{equation}
where $\Delta t$ is the sampling interval.

Unfortunately, accuracy cannot be guaranteed when depending on accelerometer and gyroscope because of the noise presence and the gyroscope drift. Thus, the magnetometer is involved to determine the yaw angle rotation which improves the gyroscope drift. However, readings from the magnetometer sensor can be affected by surrounding metals or electronic objects \cite{zhu2004real}. Using IMUs, the data for calibration is collected only in the gravitational field so that the calculated scale factor of each axis is limited to the range of [-1g , 1g]. Such limitation prevents tasks with large gravity value \cite{cai2013accelerometer,zhang2021low}. However, this limitation has been overcome by rotating the accelerometer sensor around a fixed point to detect centripetal and Euler acceleration. As a result, the detection range of the accelerometer sensor has been expanded \cite{sohrabi2017accuracy, olivares2009high, kozlov2014imu, aktakka2017chip}.    

\subsection{Data Imputation}
 Analyzing datasets is vital since they are considered a rich source of information for different type of knowledge. Nevertheless, missing data in these datasets can prevent the generation of complete information to make a wise intelligent decision \cite{adhikari2022comprehensive}. Hence, the issues that arise when data are missing can be bias and lack of recover-ability. For this reason, several studies have been conducted on the missing data imputation to impute the missing data \cite{adhikari2022comprehensive,hasan2021missing,luo2022evaluating,singh2022efficient}. The data imputation techniques aim to replace any missing data samples by randomly estimating data from the same datasets. The data imputation can undergo single or multiple imputation. In single imputation, only one estimate is used, whereas various estimates are used in multiple imputation. The missing data can be categorized as 1) missing completely at random, 2) missing at random, and 3) missing not at random \cite{donders2006gentle,lin2020missing}. 

The complete missed data at random arises when parts of the data collected are missing by design due to out-of-hand circumstances (i.e., unobserved data). In other words, missingness occurs during the data acquisition process and thus becomes of no interest. Hence, there is no correlation between the observed and missing data. A popular example is data being missed while using IMUs due to sensor failure or network connection problems. In this case, there is no bias to be introduced and data is estimated from the observed original data on average. The standard estimates errors are usually large due to the reduced sample size \cite{little2014joys,leurent2018sensitivity,de2013multiple}. Missing data undergoes missing at random if the missing information is from the observed data after confirming the dataset. Thus, correlation exists between the observed and missing data  \cite{adhikari2022comprehensive}. 

In both cases, the multiple imputation are utilized to replace the missing data with appropriate predictive one. The idea behind the multiple imputation is to predict the missing values using the observed dataset. The imputed values are estimated rather than known or uncertain, this process is repeated several times to create several complete datasets. The analysis model is then fitted to each generated dataset, and the results are combined for inference using Rubin’s MI rules \cite{leurent2018sensitivity, rubin2004multiple}. However, this estimation remains inefficient enough since the data still incomplete. The third technique occurs when valuable information is lost from the dataset and there is no universal method to handle the missing data properly. Thus, the missing data depends on its value. Hence, missing data follows missing not at random when it is not classified as missing at random or missing complete at random \cite{hasan2021missing, lin2020missing}. 
  
\subsection{Single Imputation Methods}
Single imputation (SI) techniques are common in handling missing data in research by replacing the missing data with a single value. This includes implementing mean imputation \cite{kang2014assessing}, Last Observation Carried Forward (LOCF) \cite{wood2005comparison}, Regression Imputation \cite{jekauc2012missing}, Hot Deck Imputation \cite{wang2020implementing}, Forward Imputation \cite{solaro2018simulation}, Single Ratio Imputation \cite{takahashi2017multiple},  median imputation \cite{jang2020deep}, and $K$-Nearest Neighbors (KNN) imputation \cite{singh2024integrated}. 

SI approaches are effective in handling missing data across various sports science scenarios. Team mean imputation is used to impute workload data in youth basketball \cite{benson2021evaluating}. The average workload of the team for the specific session is used to impute missing workload data including jumps per hour and RPE for high school basketball players. However, this approach cannot always be calculated, may not account for individual variations, and could introduce bias if team performance varies significantly. KNN imputation is employed to predict missing split times for runners who did not finish the Boston Marathon \cite{hammerling2014completing}. Local regression based on KNN is used to estimate missing times. However, performance depends on the choice of neighbors and may not generalize well to different datasets. Last Observation Carried Forward (LOCF) is used in \cite{kato2025relationship} to monitor Total Quality Recovery (TQR) scores and race performance for college swimmers over two seasons. The missing TQR score is imputed with the TQR score recorded for that same participant on the preceding day. Although LOCF is generally a less sophisticated imputation approach, it unrealistically assumes that a variable remains constant over the period where data is missing. Therefore, LOCF would not be able to capture if the recovery status of an athlete change daily. 

In addition, when model-based imputational methods are inappropriate, due to limitations present in the data, single-imputation procedures, such as random hot-deck imputation, can be an alternative. Random hot-deck imputation is a useful procedure because it yields plausible imputed values through matching records that contain missing data to records with complete observations, and hence does not arbitrarily alter the dataset \cite{wang2020implementing}. SI procedures can also be useful for analyzing changes in outcome variables such as behaviors, performance and psychological constructs, and is useful for preserving survey continuity when evaluating intended changes resulting from interventions \cite{barker2011single}. Single imputation can be a simple way to impute missing values when gaps are short or missing source assignment lengths are very short \cite{ramli2013roles}. Finally, Forward Imputation and missForest can potentially also yield robust imputation results across commonly encountered data patterns, both of which exhibit varying degrees of excess kurtosis, skewness and correlated structures. Forward Imputation and missForest are useful when timely and appropriate imputation is required, without strong distributional assumptions \cite{solaro2018simulation}.

Although SI techniques are often used to handle missing data, there are several important issues related to validity and reliability with the use of SI. SI techniques will replace each missing value with a single value. When replacing a missing value with a single, fixed value, SI techniques do not take into account the imputation uncertainty with missing data. Risking to cause an under-estimation of standard errors thus leading to confidence intervals that are overly narrow, which may result in an increased rate of false positives \cite{donders2006gentle}. Replacing a missing value with a single value can also affect the relationships that exist; it can distort the correlational relationships that we are investigating, as well as the coefficients we are testing. For example, it is known that mean imputation typically reduce correlations. This is challenging within sports science where understanding the relationships between performance metrics, training loads, and physiological responses. The validity of single imputation methods will depend upon the imputation methods chosen. Multiple different SI methods can produce very different results and thus impact the reliability and consistency of the findings, leading to biased imputations \cite{benson2021evaluating}. Mean imputation can distort the distribution of the data.

Moreover, data in sports science can be highly variable and context-specific where single imputation methods may not work well with complex data structures. This can lead to implausible imputed values and further bias \cite{wang2022implementing}. Single imputation methods are also less effective when data are not missing completely at random MCAR. They do not perform well under missing at random (MAR) or missing not at random (MNAR) conditions, which are common in sports science data. Methods including mean, median, or mode imputation often fail to account for the underlying mechanisms that cause the data to be missing, leading to biased estimates and incorrect conclusions \cite{Ette2006245, ramli2013roles}. Single imputation is also limited in longitudinal sports science related studies. This because it can lead to biased estimates of trends and associations over time where tracking changes in performance or health metrics over time is crucial \cite{benson2021evaluating}. For instance, it does not adequately handle the correlation between repeated measures, which can distort the analysis of longitudinal data \cite{wood2005comparison}.

\subsection{Multiple Imputation Methods}

Multiple Imputation (MI) is an a statistical technique used to handle missing data, MI generates multiple datasets with imputed missing data, analyzes each one separately, and combines results to represent uncertainty due to missing data \cite{ser2016performance}. The primary strength of MI over SI methods is MI estimates the parameters with lesslimits of bias, and more accurately derives variance estimates and confidence intervals \cite{sterne2009multiple}. Methods for using MI that have been seen in sport science include, chained equations \cite{sterne2009multiple}, joint model\cite{mistler2017comparison}, markov chain bootstrap \cite{zhang2004nonparametric}, monotone imputation \cite{white2010avoiding}, and random hot deck \cite{wang2020implementing}.

MI methods are already routinely and commonly used in sport science. A recent example of MI can be seen in the recent article evaluating the monitoring of athlete workloads \cite{benson2021evaluating}, for longitudinal studies that monitor athlete workload. The authors proposed to impute workload for youth basketball players on the workload variable of jumps per hour, and used regression based methods in MI that used numerous predictors. The authors used MI to only impute the workload variables, and did not include any other non-workload data that might have improved their imputations' quality. In sport injury epidemiology, a predictive model-based MI is used to estimate missing weekly game hours for 2098 youth ice hockey players \cite{kang2014assessing}. The imputed estimates represent the mean hour estimates from the imputed samples. The statistical models used for their data such as included Poisson, zero-inflated poisson, and negative binomial regressions on the imputed datasets to estimate injury rate ratios. In \cite{kang2014assessing}, they found that when the dataset had less to moderate proportional missing data for weekly game hours, MI performed had a similar performance to mean imputation.

A new MI method for auto-correlated multivariate count data from accelerometers was developed in \cite{ae2018missing}. It uses mixture of zero-inflated Poisson and Log-normal to handle characteristics including autocorrelation and over/under-dispersion of count data. However, the characteristics of accelerometer data such as autocorrelated, and multivariate counts are still challenging for imputation. A framework for multilevel data was proposed in \cite{nguyen2024filling} to estimate aging curves for offensive players in Major League Baseball (MLB). This study treats unobserved seasons due to player dropouts as missing data. Player performance metrics associated with missing seasons are imputed and aging curves are then constructed based on these imputed datasets. The main limitation is accurately modeling the reasons for player dropout and the performance trajectories of those who dropout.

Despite the advantages of using MI over SI, MI has several limitations. MI typically performs best under MCAR or MAR assumptions. If data are MNAR, standard MI may produce biased results unless the MNAR mechanism is explicitly modeled or sensitivity analyses are conducted \cite{woods2024best}. Moreover, implementing MI methods correctly requires statistical knowledge and familiarity with relevant software \cite{benson2021evaluating}. The accuracy of MI methods is also challenging due to its heavily dependence on the correct specification of the imputation model. This includes choosing the right variables to include in the model and specifying the appropriate relationships \cite{li2015multiple}. Another limitation is that MI methods can be computationally intensive for large datasets with many variables and complex patterns of missingness \cite{woods2024best}. However, modern software and multi-core processors have alleviated this to some extent.

\subsection{Gap Analysis}

MoCap systems utilizing IMUs have become essential as important analysis tools for studying human activities and sports across multiple fields: sports science, human performance assessment, diagnostic properties of certain neurodegenerative disorders, and 3D animation \cite{van2018accuracy,jalata2021movement}. MoCap systems have a high utility for human activity analysis, however, there are multiple potential sources of error in the IMU sensor data when using these systems: battery limitations, network issues, magnetic fields, drift and placement variability, which can cause presence of missing data  (i.e., missing time-series data) \cite{yahyamotion}. It is critical that any lost or missing data have to be addressed  to eliminate bias and inaccurate information that restrict data reliability for subsequent analyses and decision-making, and to assess how various data sources interface. Therefore, effective and appropriate data imputation methodologies should be explored and applied for use in MoCap analysis.
The literature supports many data imputation methods but focuses on SI and MI methods, and some basic statistical or non-statistical methodologies, however only discusses SI or MI methodologies in relation to sports science broadly; thus presenting a notable gap in literature for the classification and evaluation of data imputation methodologies specific to MoCap data and IMU wearable sensors for enhacing capabilities of MoCap \cite{yahyamotion}. Entirely separate from MoCap data, data imputation studies focus on other types of data with variability in missing data mechanisms and contexts, that are not reflective of continuous time series data, or the nature of the missing time series data (e.g., Sequential, biomechanical constraints). In our review of the literature, it reveals a couple of important gaps in the study that this study addresses, respectively:

\begin{enumerate}
    \item \textbf{Lack of comprehensive performance evaluation for IMU-based MoCap imputation:} To date, no research has systematically examined and compared the performance of well-established data imputation strategies, including ML, DL, and statistical methods, specifically for MoCap data acquired from IMU sensors. This absence limits the ability of researchers and practitioners to select the most appropriate imputation technique for specific MoCap missing data scenarios, hindering accurate data reconstruction and subsequent analysis.
    \item \textbf{Absence of a standardized MoCap benchmark dataset for imputation:} A crucial prerequisite for reproducible research and comparative analysis is the availability of a standardized dataset. Currently, there is no publicly available MoCap dataset explicitly designed and proposed for benchmarking the performance of various data imputation methods
. This hinders the consistent evaluation and advancement of imputation techniques tailored to the unique complexities of MoCap data.
\end{enumerate}

This research directly addresses these important gaps by providing the first thorough performance comparison of the major data imputation methods in the context of ML, DL and statistical methods as it relates to IMU based MoCap data. As a result of the comparison, we offer definitive recommendations for what are the best techniques for specific cases of incomplete MoCap data. Furthermore, we have introduced the first dedicated MoCap dataset as a benchmark for research focused on missing value imputation, to stimulate future research and ensure that there is a consistent method of benchmarking use in the community. By addressing these gaps, this work contributes to theories of data processing and advances the state-of-the-art of robust information fusion from IMU based motion data. We also offer practical, official recommendations for best practices in enhancing data quality and use.

%-------- Mahmoud ---------

\section{Methodology}
\label{sec:methodology}
%%%%%%%%%%%%%%%%%%%%%%%%%%%%%
This section details the experimental methods used to test different data imputation methods on a multivariate time-series dataset containing simulated human motion data. We describe the dataset's structure, the methodology for simulating missing data under a variety of mechanisms, the imputation framework, including which specific algorithms were tested, and the measures used to evaluate the performance of the imputation algorithms. 

\subsection{Overview}

\begin{figure}
\centering  
    \includegraphics[width=0.4\linewidth]{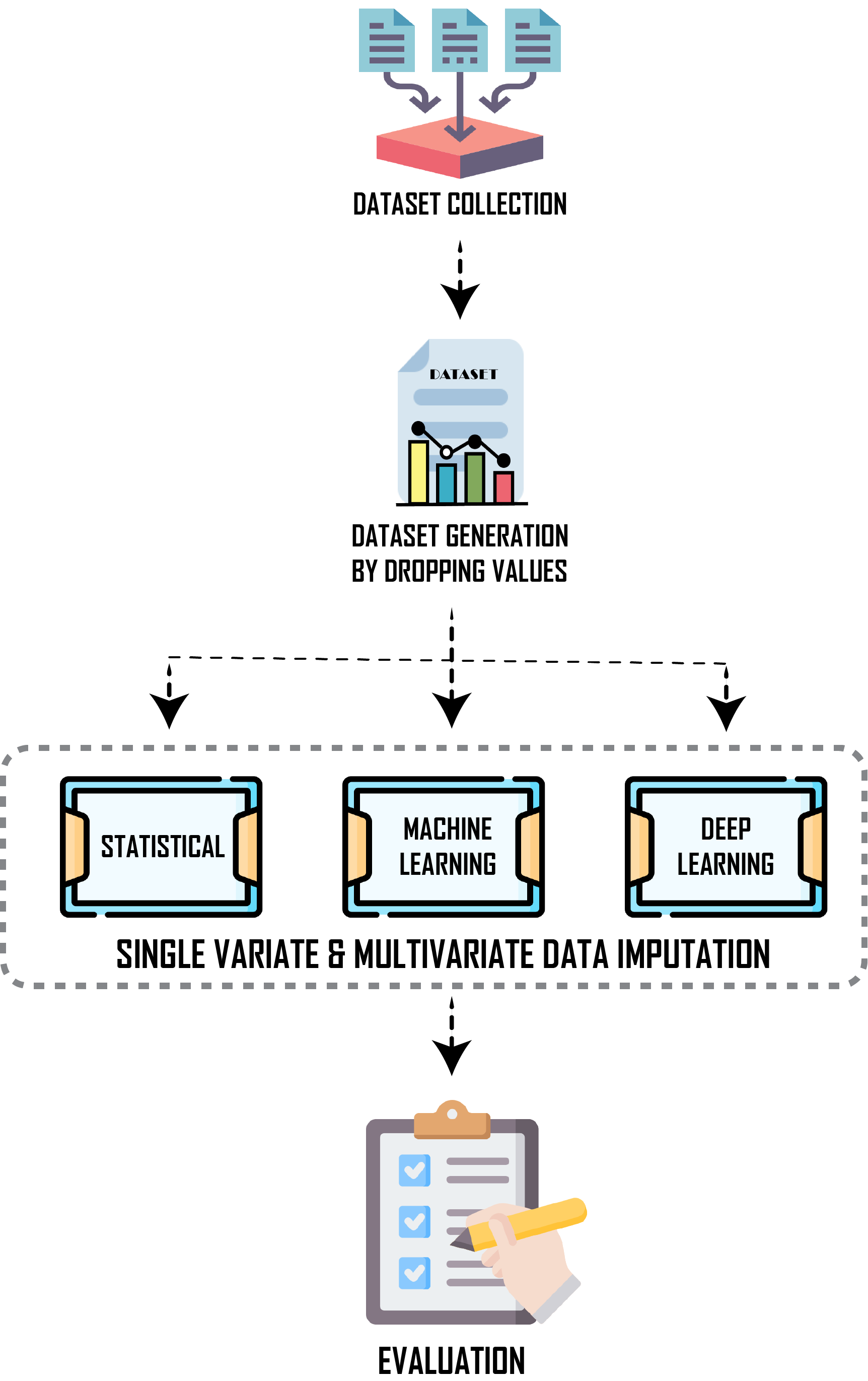}          
  \caption{The proposed MoCap data imputation methodology.\label{fig:overview}}
\end{figure}

The proposed methodology for evaluating the MoCap data imputation is depicted in Fig. \ref{fig:overview}. In order to simulate real-world scenarios, the procedure begins with the entry of MoCap data with purposefully absent values. This raw data is cleaned, normalized, and standardized in order to prepare it for the succeeding imputation phase.

Once the data has been preprocessed, it is forwarded to the basis of our system: a suite of imputation algorithms. This suite contains a variety of methods, including ML strategies such as Bsi-ML, Iterative Imputer-ML, and KNN-ML, the DL method DL-based Generative Adversarial Imputation Network (DL-GAIN), and statistical methods such as Simplefill mean and Simplefill median. By employing a variety of techniques, our proposed study aims to accommodate the diverse details and complexities inherent in MoCap data, thereby guaranteeing optimal data imputation.

The processed data are subjected to a strict evaluation after imputation. This phase of evaluation examines the effectiveness of the selected imputation algorithms by comparing the imputed values to the original data. Metrics calculation, including the critical Mean Absolute Error (MAE), is employed to further refine the evaluation. These metrics quantify the accuracy and dependability of the imputed data, demonstrating the model's proficiency.

After an effective evaluation and calculation of metrics, the system produces the imputed dataset. This dataset, augmented by the systematic processes of our methodology, not only represents the missing MoCap data but also demonstrates the convergence of ML, DL, and statistical techniques in addressing the challenges posed by missing data. Our proposed system offers an effective solution for MoCap data imputation with this integrated approach.

\subsection{Dataset and Preprocessing}
\label{subsec:dataset}

Due to the nature of our study, we utilize a multivariate time-series dataset with \( P \) distinct persons (players) who are performing a specific skill which is publicly available \cite{fathalla2023real}. The utilized dataset is produced by collecting data from IMUs from 53 karate players performing four different skills \cite{fathalla2023real}.   Each player has data collected for \( T \) discrete time points and \( A \) features that correspond to measured angles or kinematic variables. The original complete dataset is stored as a tensor  \( \mathbf{X} \in \mathbb{R}^{P \times T \times A} \). For the experiments we describe in this paper, we have a dataset with dimensions \( P=53 \), \( T=100 \), and \( A=39 \). While this dataset does not suffer from missing values, we proposed using it to artificially include missing values with three different mechanisms.

We proposed a preprocessing step to the original dataset which is normalization. The dataset undergoes min-max normalization prior to the use of some of the imputation methods, specifically those that are especially sensitive to feature scaling such as the neural network-based methods like GAIN. For each feature \( a \)  (angle), we scale the time series  \( \mathbf{X}_{:,:,a} \) across all players and time points to the range [0, 1] using the following formula: 

\begin{equation}
    \mathbf{X}_{\text{norm}, p,t,a} = \frac{X_{p,t,a} - \min_{p',t'}(X_{p',t',a})}{\max_{p'',t''}(X_{p'',t'',a}) - \min_{p',t'}(X_{p',t',a}) + \epsilon}
    \label{eq:normalization}
\end{equation}
where the $min$ and $max$ are calculated over the observed entries only for that feature \( a \), and  \( \epsilon \) is a small constant (e.g., \(10^{-6}\)) for numerical stability, and $p'$, $p''$, $t'$, and $t''$ are dummy indices that range over all players and time points, respectively. We take the parameters \( \min(\cdot) \) and \( \max(\cdot) \) for each feature and save them in order to perform an inverse transformation (renormalization) back to the original scale of the data post imputation so that we can estimate meaningful error.

\subsection{Simulation of Missing Data}
\label{subsec:missing_data_simulation}

\begin{figure}[H]
        \subfloat[Random missing data\label{fig:sample_random_points}]{
            \includegraphics[width=0.5\linewidth]{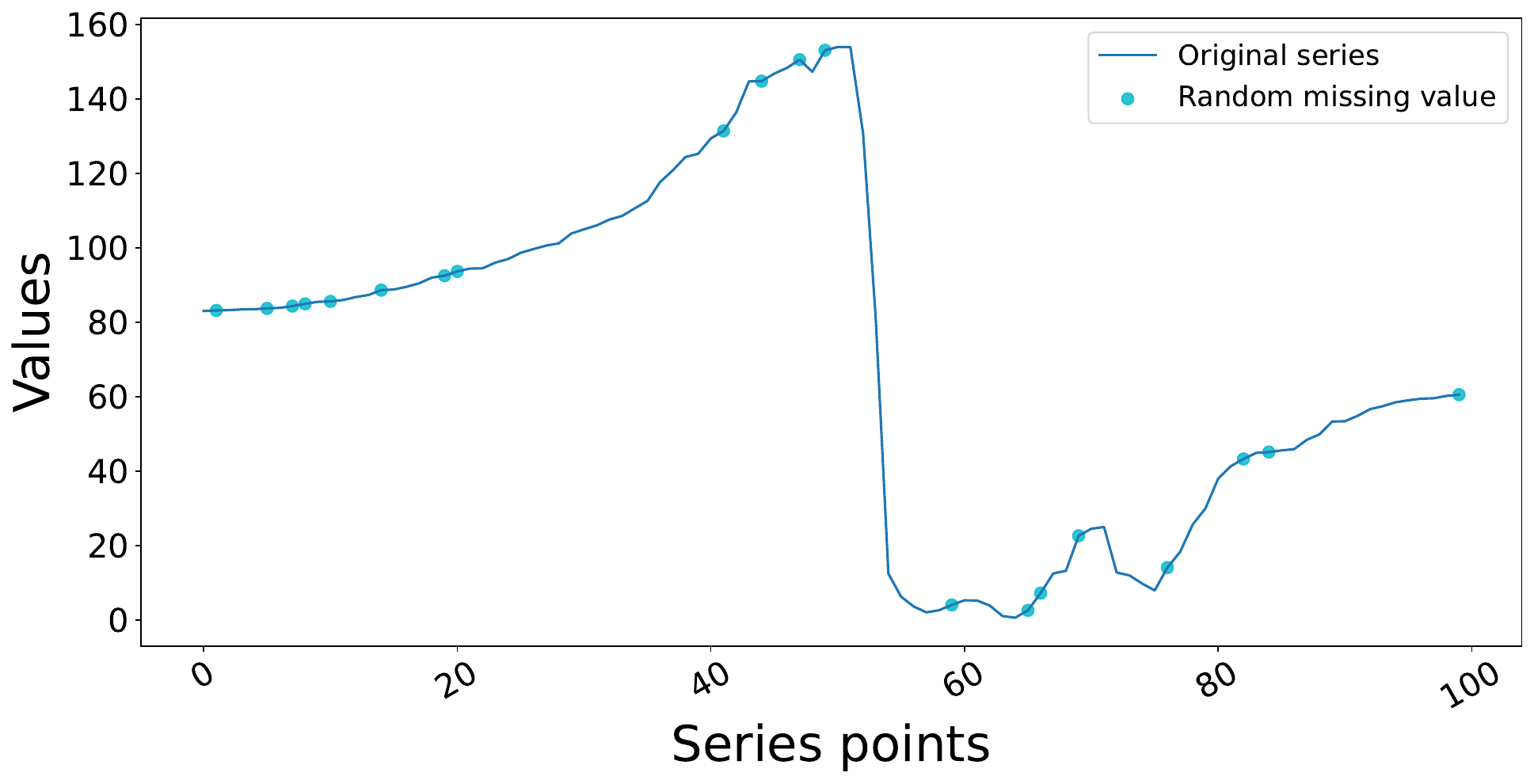}}
~
		\subfloat[Generating missing data of transition and random points\label{fig:sample_transition_points}]{
            \includegraphics[width=0.5\linewidth]{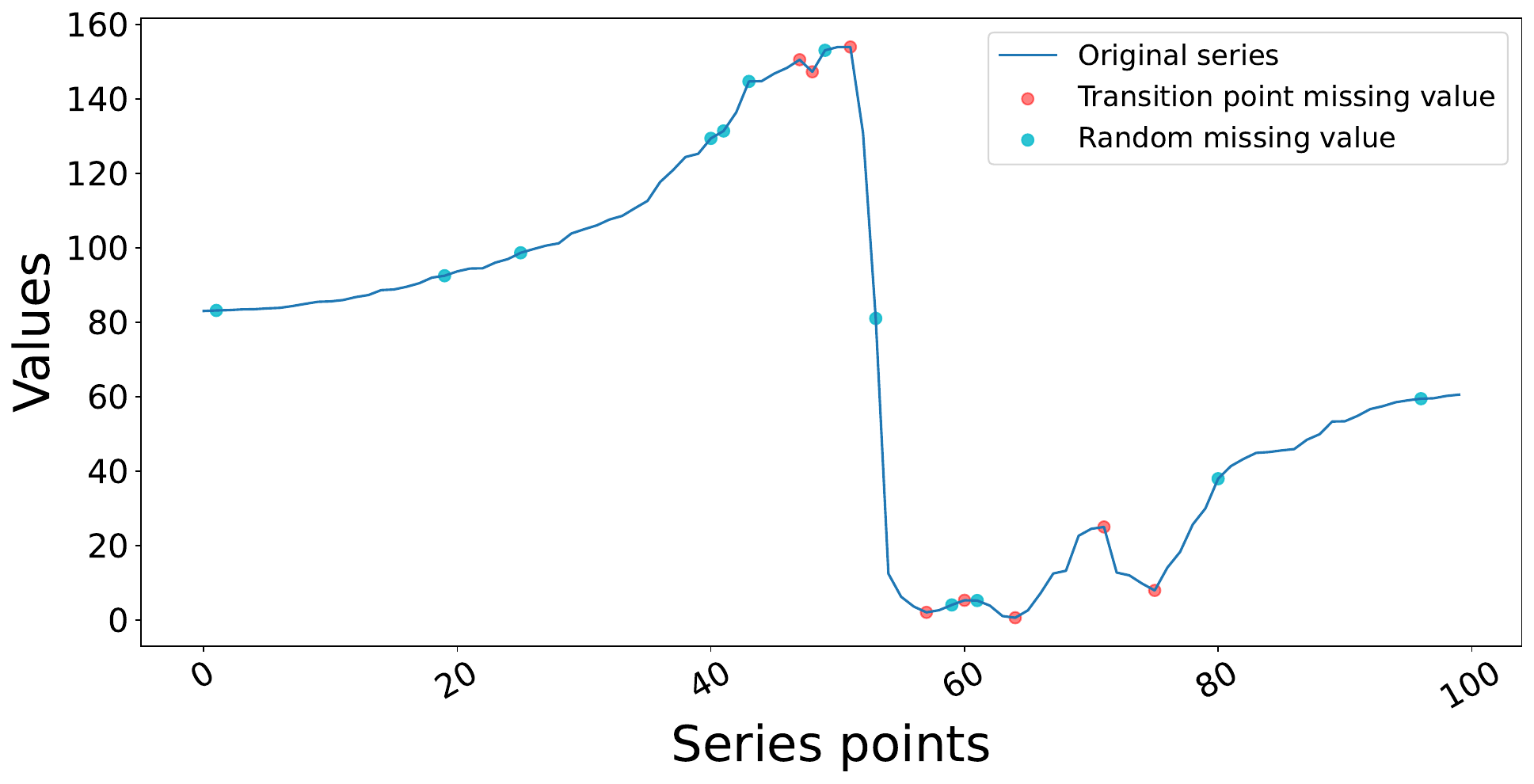}}

        \centering
        \subfloat[Generating of continuous intervals of missing data\label{fig:sample_interval}]{
            \includegraphics[width=0.5\linewidth]{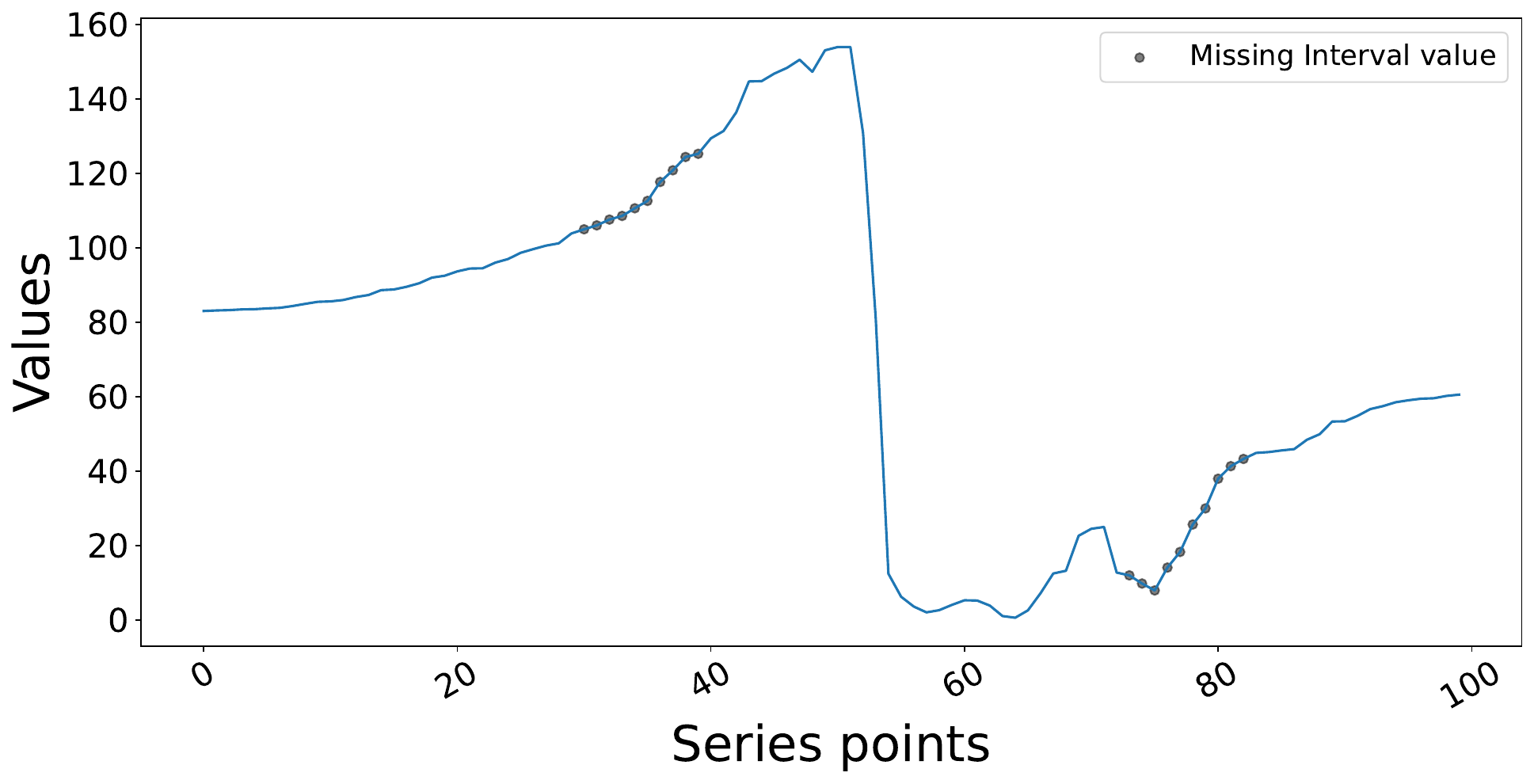}}
    \caption{Samples of the proposed/utilized three methods for generating 20 missing data points for the same time series.\label{fig:SK_error}}
\end{figure}  

Missing values were synthetically generated within the complete dataset \( \mathbf{X} \) to create controlled experimental conditions for assessing imputation performance. This process produces a missing data tensor $X_{\text{miss}}$, and then a corresponding binary mask tensor \( \mathbf{M} \in \{0, 1\}^{P \times T \times A} \). The symbol \( M_{p,t,a} = 1 \) indicates the value \( X_{p,t,a} \) is missing (NaN in  \( \mathbf{X}_{\text{miss}} \)), while \( M_{p,t,a} = 0 \) signifies that the value was observed. 

The missing data entries were independently generated for each univariate time series  \( \mathbf{X}_{p,:,a} \) based on a specified number of missingness count \( k \), corresponding to a fraction \( k/T \) of the total series length (which varied between 5\% and 30\% in our experiments), and based on one of three different missing data mechanisms, denoted by \( \zeta \): 

\begin{enumerate}
 
    \item \textbf{Missing Completely At Random (MCAR):} From  \( \{0, 1, ..., T-1\} \), \( k \) distinct time indices were sampled uniformly without replacement for each series \( \mathbf{X}_{p,:,a} \), and the corresponding entries in \( \mathbf{M}_{p,:,a} \) were assigned a value of 1. This mechanism assumes that missingness was completely independent of observed and unobserved values. A sample of this approach is depicted in Fig. \ref{fig:sample_random_points}.

    \item \textbf{Value-Dependent Missingness at Transition Points:} This mechanism simulates a form of value-dependent missingness, where the probability of data loss is intentionally correlated with the local dynamics of the signal itself. Specifically, we target transition points (local minima and maxima), as these points of high kinetic change can be more susceptible to measurement error or signal clipping in real-world MoCap applications. First, for each time series $X_{p,:,a}$, we identify the complete set of transition point indices $T_{p,a}$. Then, we randomly sample $k' = \min(k, |T_{p,a}|)$ indices from these transition points. If the total number of missing points to be introduced, $k$, is greater than the number of available transition points, we sample the remaining $k - k'$ indices from the non-transition points. All sampled indices were set to missing ($M_{p,t,a} = 1$). Figure~\ref{fig:sample_transition_points} shows a sample of this approach.

    \item \textbf{Block Missingness (Structured):} we introduced blocks of contiguous missing values. For the purposes of sampling blocks, the total length of the time series \( T \) was notionally divided into segments of \( N_b \). For each segment we placed a block of a predetermined size \( L_b \) (where \( L_b \) was identified to approximate the total count of \( k \)  across all blocks) starting from a random index within the limits of the segment. Again, all sample indices in these  \( N_b \) blocks were converted to missing in \( \mathbf{M} \). In the case of overlaps in blocks or deviations in total segment length which deemed the total count < \( k \) (or > \( k \)) we made adjustments. Fig. \ref{fig:sample_interval} depicts a sample of this dataset. 
\end{enumerate}
The missing data mask \( \mathbf{M} \) was generated according to these mechanisms as formalized in Algorithm ~\ref{alg:mask_generation} The algorithm iterates through each player \( p \) and angle \( a \), applies the missingness mechanism \( \zeta \) to the corresponding indexed time series to identify the set of indices 'idx' to undergo masking, and updates the mask tensor M, specifying clarity on what mechanisms' logic was used for each mechanism (MCAR, Transition, Block) in the specified conditional statements.

\begin{algorithm}
\caption{GenerateMissingMask($X, k, \text{mechanism}$)}
\label{alg:mask_generation}
\begin{algorithmic}[1]
\Require Complete data tensor $X \in \mathbb{R}^{P \times T \times A}$, Missing count $k$, \( \zeta \) $\text{mechanism}$
\Ensure Mask tensor $M \in \{0,1\}^{P \times T \times A}$
\State Initialize $M \leftarrow \mathbf{0}^{P \times T \times A}$
\For{$p \in \{1, \dots, P\}$}
    \For{$a \in \{1, \dots, A\}$}
        \State Let $S \leftarrow X_{p,:,a}$ be the time-series
        \State Let $I_{miss} \leftarrow \emptyset$ be the set of indices to mask

        \If{$\text{mechanism} = \text{MCAR}$}
            \State $I_{miss} \leftarrow \text{RandomSample}(\{1, \dots, T\}, k)$
        
        \ElsIf{$\text{mechanism} = \text{Transition}$}
            \State $I_{trans} \leftarrow \text{FindLocalExtremaIndices}(S)$
            \State $I_{sample} \leftarrow \text{RandomSample}(I_{trans}, \min(k, |I_{trans}|))$
            \If{$|I_{sample}| < k$}
                \State $I_{remain} \leftarrow \{1, \dots, T\} \setminus I_{trans}$
                \State $I_{sample} \leftarrow I_{sample} \cup \text{RandomSample}(I_{remain}, k - |I_{sample}|)$
            \EndIf
            \State $I_{miss} \leftarrow I_{sample}$

        \ElsIf{$\text{mechanism} = \text{Block}$}
            \State Let $N_b, L_b$ be block parameters s.t. $N_b \cdot L_b \approx k$
            \State Partition $\{1, \dots, T\}$ into $N_b$ segments, $Seg_1, \dots, Seg_{N_b}$
            \For{$i \in \{1, \dots, N_b\}$}
                \State $start\_idx \leftarrow \text{RandomInt}(\min(Seg_i), \max(Seg_i) - L_b)$
                \State $I_{miss} \leftarrow I_{miss} \cup \{start\_idx, \dots, start\_idx + L_b - 1\}$
            \EndFor
            \State Trim or extend $I_{miss}$ to ensure $|I_{miss}| = k$
        \EndIf
        
        \State $M_{p, I_{miss}, a} \leftarrow 1$
    \EndFor
\EndFor
\State \textbf{return} $M$
\end{algorithmic}
\end{algorithm}

\subsection{Imputation Framework Contexts}
\label{subsec:imputation_contexts}

To clarify on how different imputation methods can take advantage of the structure in the dataset, we applied algorithms in three different information contexts as follows:

\begin{enumerate}
    \item \textbf{Univariate Context:} 
    Here data is completed independently for each individual time series \( \mathbf{X}_{p,:,a} \in \mathbb{R}^T \). Univariate algorithms will only learn from information in the individual series being completed, a limitation of univariate context is that we ignore any potential relationships across players or angles. Input data is treated as a vector with length \( T \).

    \item \textbf{Multivariate Context (Across Players):} 
    For a fixed angle \( a \) we conceptualize imputation as the data matrix \( \mathbf{X}_{:,:,a} \in \mathbb{R}^{P \times T} \) (or the transpose). As a result algorithms that in the Multivariate context can model correlations, or similarities, across different players time series across angles will take advantage of information from cohort.
   
    \item \textbf{Multivariate Context (Across Angles):} 
    For a fixed player \( p \), we represent imputation as the data matrix \( \mathbf{X}_{p,:,:} \in \mathbb{R}^{T \times A} \). Algorithms that operate in the multivariate context can take advantage of inter-feature correlations, essentially learning about how various angles (kinematic variables) are related from the same subject over time. 

\end{enumerate}

This exploration of a potentially multi-context, enables a more precise evaluation of the advantages gained or lost, by utilizing information across players or angles, compared to univariate means.

\subsection{Imputation Algorithms}
\label{subsec:algorithms}

We implemented and compared a diverse set of imputation algorithms, spanning statistical baselines to state-of-the-art deep learning models:

\begin{itemize}
    \item \textbf{Statistical Baselines:} Simple, computationally inexpensive methods including Mean, Median, and Random Sample imputation, applied within the relevant context (univariate series, or multivariate scope across players/angles for calculating the statistic or sampling pool).

    \item \textbf{Classical Machine Learning Methods:} Algorithms primarily sourced from the \texttt{fancyimpute} library:
        \begin{itemize}
            \item \textit{KNN:} Estimates missing values using a weighted average of the values from the \( K \) most similar samples (neighbors), based on observed features.
            \item \textit{Matrix Factorization (SoftImpute, IterativeSVD):} These methods approximate the data matrix with a low-rank factorization, effectively filling missing entries based on learned latent factors. SoftImpute uses nuclear norm regularization, while IterativeSVD employs truncated SVD iteratively.
            \item \textit{IterativeImputer:} Models each feature with missing values as a function of other features using a regression model (e.g., Bayesian Ridge). It iteratively predicts and updates missing values until convergence.
              \item \textit{Optimal Transport Imputation (OT) \citep{muzellec2020missing}:} This method leverages Optimal Transport theory, specifically minimizing the Sinkhorn divergence \( S_{\epsilon}(\cdot, \cdot) \) between empirical distributions formed by batches of data. We utilized the \texttt{BatchSinkhornImputation} approach, where the missing values themselves are treated as learnable parameters \( \theta \). In the following text, we will call this method BSI-LM. These parameters are optimized by minimizing the expected Sinkhorn divergence between pairs of randomly drawn mini-batches \( (\mathcal{B}_1, \mathcal{B}_2) \) from the currently filled dataset \( \mathbf{X}_{\text{filled}}(\theta) \):
            \begin{equation}
                \min_{\theta} \mathcal{L}_{\text{OT}} = \mathbb{E}_{(\mathcal{B}_1, \mathcal{B}_2)} [ S_{\epsilon}(\mathbf{X}_{\text{filled}}(\theta)[\mathcal{B}_1], \mathbf{X}_{\text{filled}}(\theta)[\mathcal{B}_2]) ]
                \label{eq:ot_loss}
            \end{equation}
            The expectation is approximated using \( N_{\text{pairs}} \) samples per gradient step. 
        \end{itemize}

    \item \textbf{Deep Learning Methods:}
        \begin{itemize}
            \item \textit{GAIN \citep{Yoon2018GAIN}:} GAIN employs a minimax game between a Generator (\( G \)) and a Discriminator (\( D \)). The Generator ($G$) attempts to impute the missing values in the data tensor $X_{\text{miss}}$ given a mask $M$ and a noise tensor $Z$, producing the final imputed tensor $\hat{X} = X_{\text{miss}} \odot (1-M) + G(X_{\text{miss}}, M, Z) \odot M$. The discriminator $D$ tries to distinguish observed components from imputed ones based on \( \hat{\mathbf{X}} \) and a hint vector \( \mathbf{H} \) (partially revealing \( \mathbf{M} \)). The objectives are:
            \begin{align}
                \min_{D} \mathcal{L}_D &= -\mathbb{E}_{\mathbf{X}, \mathbf{M}, \mathbf{H}} [\mathbf{M} \log D(\hat{\mathbf{X}}, \mathbf{H}) + (\mathbf{1}-\mathbf{M}) \log(1 - D(\hat{\mathbf{X}}, \mathbf{H}))] \\
                \min_{G} \mathcal{L}_G &= \mathbb{E}_{\mathbf{X}, \mathbf{M}, \mathbf{H}} [-\log(D(\hat{\mathbf{X}}, \mathbf{H}))] + \alpha \mathbb{E}_{\mathbf{X}, \mathbf{M}} [\| \mathbf{M} \odot (\mathbf{X} - G(\mathbf{X}, \mathbf{M}, \mathbf{Z})) \|_2^2]
            \end{align}
            where \( \odot \) denotes element-wise multiplication and \( \alpha \) is a hyperparameter balancing the adversarial loss and a direct reconstruction (MSE) loss on observed components. 
        \end{itemize}
\end{itemize}

\subsection{Evaluation Metrics}
\label{subsec:evaluation}
To quantitively assess imputation accuracy, values \( \hat{\mathbf{X}}_{p,t,a} \)  were compared to their respective known ground truth values \( \mathbf{X}_{p,t,a} \) only at the locations where the data had been artificially made missing (i.e., \( M_{p,t,a} = 1 \)). The primary measure of performance for our study was the Mean Absolute Error (MAE) to quantify the average size of the imputation: 

\begin{equation}
    \text{MAE} = \frac{\sum_{p,t,a} M_{p,t,a} \cdot | X_{p,t,a} - \hat{X}_{p,t,a} |}{\sum_{p,t,a} M_{p,t,a}}
    \label{eq:mae}
\end{equation}

We also calculated the standard deviation of the absolute errors, \( \text{STD}(\{|X_{p,t,a} - \hat{X}_{p,t,a}| \mid M_{p,t,a} = 1\}) \), which serves as a measure of variability or consistency across errors resulting from a particular method. The metrics were pooled and reported out based on the experiment context (e.g., across a whole context as an average across missing points, or per player/angle).

\subsection{Experimental Setup and Implementation}
\label{subsec:setup}

The entire experimental made use of essential libraries such as NumPy (for numerical tasks), Pandas (for handling data), Scikit-learn (for baseline ML models), TensorFlow v1 (for GAIN), PyTorch and GeomLoss (for OT imputation), and FancyImpute (for many classical methods). 

Due to the expansive scope of the experiments (e.g., multiple combinations of parameters: skills, missingness levels, mechanisms, imputation contexts, and algorithms) computational time was critically important. We used Python's \texttt{multiprocessing} library to run imputation tasks simultaneously across different CPU cores. This allowed us to assign independent imputation tasks (e.g., different series or matrices based on context) to different processes. We used shared memory arrays (\texttt{multiprocessing.Array}), built into helper functions to provide process-safe access, to accumulate results from concurrent processes. 

An overview of the whole experimental procedure, which included data generation, imputation in different configurations, and evaluation, is presented conceptually in Algorithm ~\ref{alg:config}. Algorithm ~\ref{alg:config} works through each of the experimental configurations; loads/generates the data and the mask, applies the algorithm within the context, performs calculations for performance, and stores the result. The ‘ApplyImputation‘ step presents a conceptual definition of executing the chosen algorithm, including any intended data slicing based on the context and possible parallelized operation.

In formalizing the fundamental imputation process across the experimental conditions, we present Algorithm \ref{alg:imputation}, which proposes context-aware imputation methods. Algorithm \ref{alg:imputation} is a dispatcher that prepares the data before passing it to an imputation function $f$. It essentially first copies the working data to a working dataset, denoted $X_{miss}$, which presents the data with missing values represented as $NaN$. Next, it systematically slices the data tensor based on the specified context $c$. In the case of Univariate, the data are iterated through in an individual time-series manner, whereas the Multivariate-Player and Multivariate-Angle formats create two-dimensional data matrices containing all the data by angle (all players) or by player (all angles), respectively. This structured context-aware approach enables each of the imputation methods to be applied to the data so that each method can take advantage of the different correlations; temporal, players, or features, that define the context in which they are being applied.

\begin{algorithm}
\caption{Experimental Configuration}
\label{alg:config}
\begin{algorithmic}[1]
\Require Dataset $X$, Players $P$, Time-series length $T$, Features $A$
\State \textbf{Define} Imputation Methods $\mathcal{F} \leftarrow \{\text{GAIN, IterativeImputer, KNN, ...}\}$
\State \textbf{Define} Missingness Mechanisms \( \zeta \) $ \leftarrow \{\text{MCAR, Transition, Block}\}$
\State \textbf{Define} Missingness Proportions $K \leftarrow \{0.05, 0.10, 0.15, 0.20, 0.25, 0.30\}$
\State \textbf{Define} Imputation Contexts $\mathcal{C} \leftarrow \{\text{Univariate, Multivariate-Player, Multivariate-Angle}\}$
\State \textbf{Initialize} Results storage $Results \leftarrow \emptyset$
\State
\ForAll{$f \in \mathcal{F}$}
    \ForAll{$m \in \zeta$}
        \ForAll{$i \in K$}
            \State Let $k \leftarrow \lfloor i \cdot T \rfloor$
            \State Generate Mask $M \leftarrow \text{GenerateMissingMask}(X, k, m)$
            \ForAll{$c \in \mathcal{C}$}
                \State $\hat{X} \leftarrow \text{ApplyImputation}(X, M, f, c)$
                \State $MAE \leftarrow \text{CalculateMAE}(X, \hat{X}, M)$
                \State $StdErr \leftarrow \text{CalculateStdErr}(X, \hat{X}, M)$
                \State $Results \leftarrow Results \cup \{(f, m, i, c): (MAE, StdErr)\}$
            \EndFor
        \EndFor
    \EndFor
\EndFor
\State \textbf{return} $Results$
\end{algorithmic}
\end{algorithm}

\begin{algorithm}
\caption{ApplyImputation($X, M, \text{method}, \text{context}$)}
\label{alg:imputation}
\begin{algorithmic}[1]
\Require Data $X$, Mask $M$, Imputation method $f$, Context $c$
\Ensure Imputed data tensor $\hat{X}$
\State $X_{miss} \leftarrow X \odot (\mathbf{1}-M) + \text{NaN} \odot M$
\State Initialize $\hat{X} \leftarrow X_{miss}$
\If{$c = \text{Univariate}$}
    \For{$p \in \{1, \dots, P\}$}
        \For{$a \in \{1, \dots, A\}$}
            \State Let $S_{miss} \leftarrow X_{miss}[p, :, a]$
            \State $S_{imputed} \leftarrow f.\text{fit\_transform}(S_{miss})$
            \State $\hat{X}[p, :, a] \leftarrow S_{imputed}$
        \EndFor
    \EndFor
\ElsIf{$c = \text{Multivariate-Player}$}
    \For{$a \in \{1, \dots, A\}$}
        \State Let $D_{miss} \leftarrow X_{miss}[:, :, a]$ (Shape $P \times T$)
        \State $D_{imputed} \leftarrow f.\text{fit\_transform}(D_{miss})$
        \State $\hat{X}[:, :, a] \leftarrow D_{imputed}$
    \EndFor
\ElsIf{$c = \text{Multivariate-Angle}$}
    \For{$p \in \{1, \dots, P\}$}
        \State Let $D_{miss} \leftarrow X_{miss}[p, :, :]$ (Shape $T \times A$)
        \State $D_{imputed} \leftarrow f.\text{fit\_transform}(D_{miss})$
        \State $\hat{X}[p, :, :] \leftarrow D_{imputed}$
    \EndFor
\EndIf
\State \textbf{return} $\hat{X}$
\end{algorithmic}
\end{algorithm}

\section{Results and Discussion\label{sec:results}}
In this section, the performance of various imputation methods is analyzed and interpreted for the experimental conditions described here. First, we examine the baseline performance in the challenging univariate context, subsequently examining what may be improved by utilizing multivariate information, both across players and across angles, for accurate imputation, particularly with complex missingness. We conclude with recommendations based on our analyses for researchers and practitioners.

\subsection{Performance and Limitations of Univariate Imputation}

When considering the univariate case, we treat each time-series independently, which is common when analyzing the performance of a unique or new skill. Therefore, our results show the limitations of this as performance is highly dependent on how the data is missing.

\begin{figure}[H]
        \subfloat[MAE for imputing randomly missing (MCAR) points.\label{fig:Single_0}]{
            \includegraphics[width=0.5\linewidth]{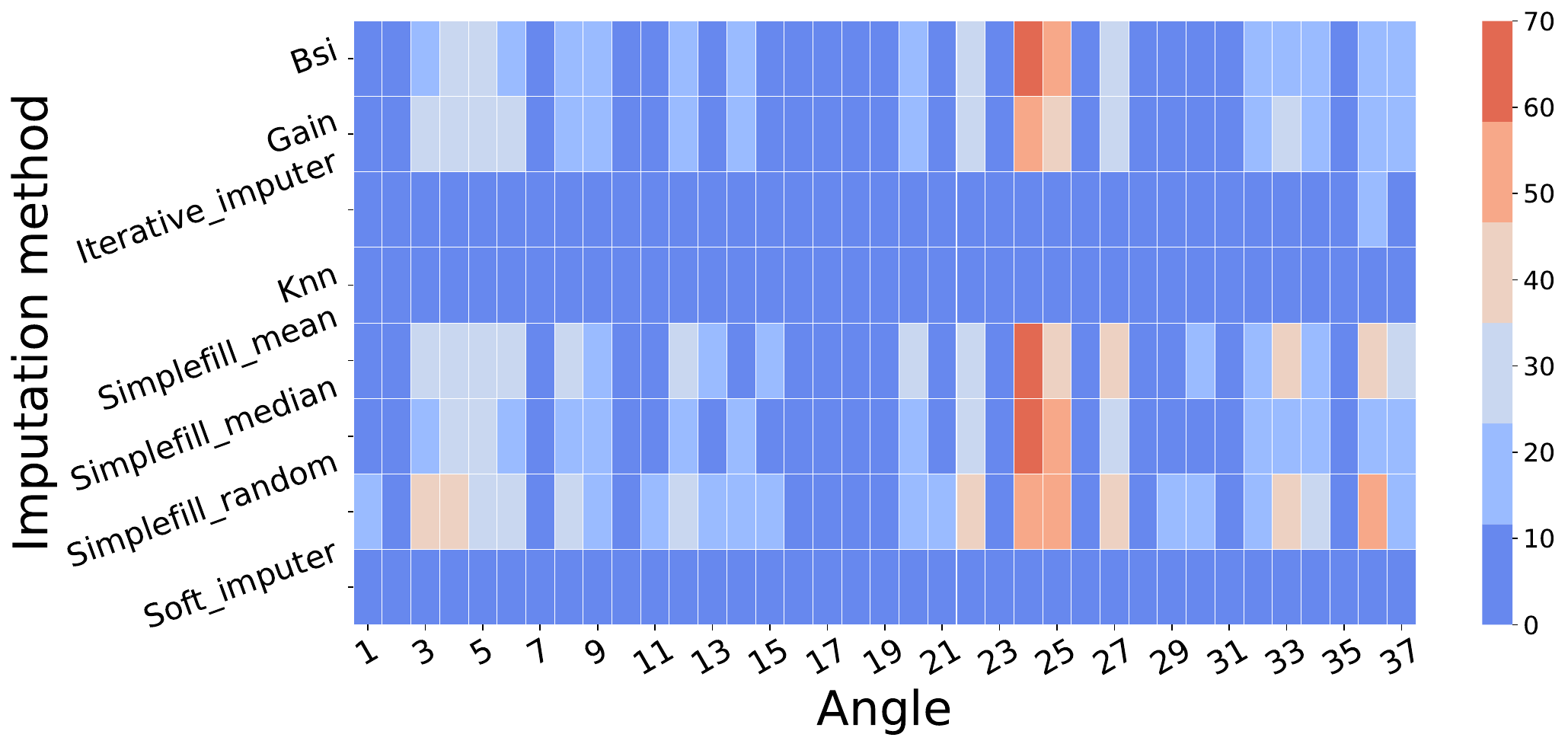}}
~
		\subfloat[MAE for imputing missing data of transition and random points.\label{fig:Single_1}]{
            \includegraphics[width=0.5\linewidth]{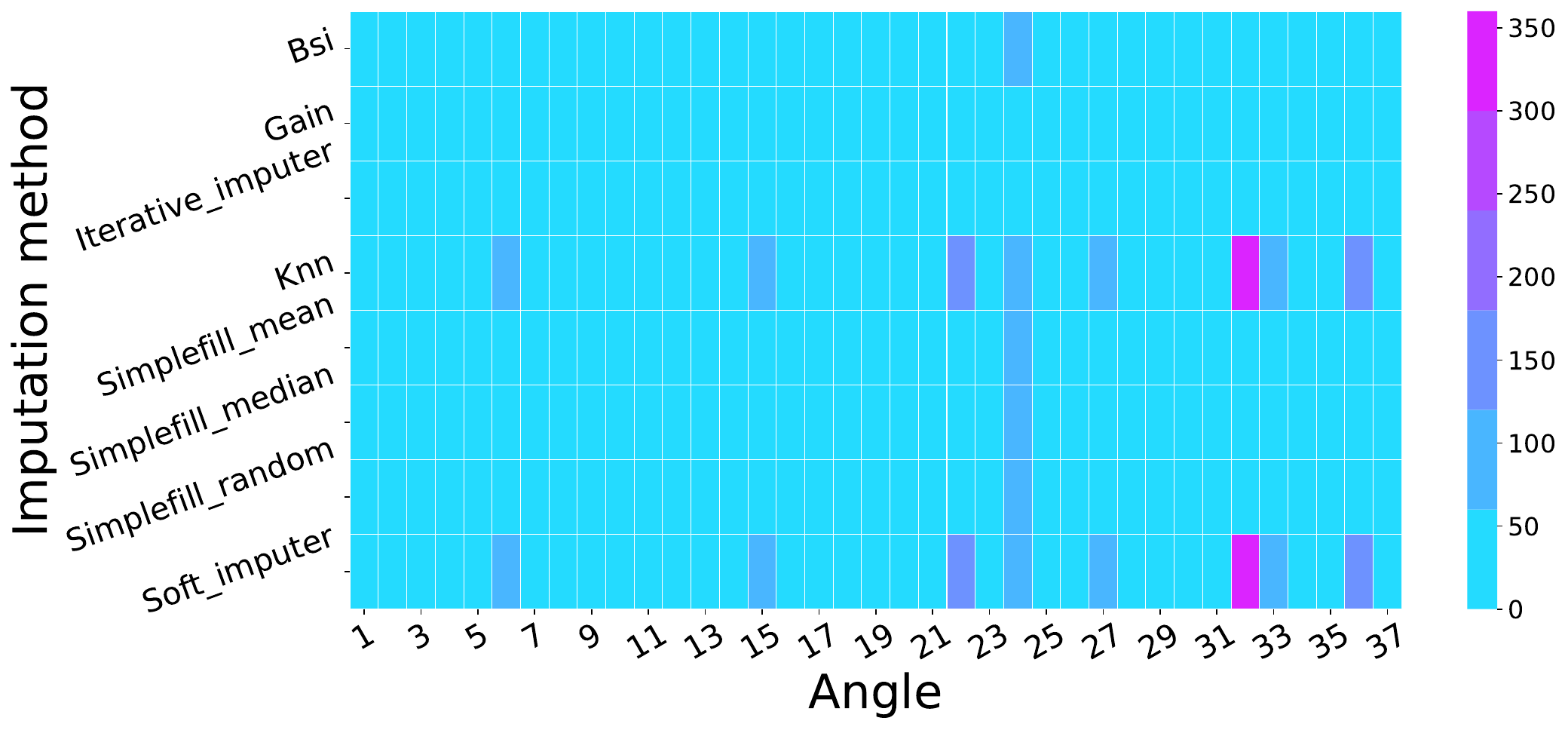}}

        \centering
        \subfloat[MAE for imputing sequences of consecutive missing data.\label{fig:Single_2}]{
            \includegraphics[width=0.5\linewidth]{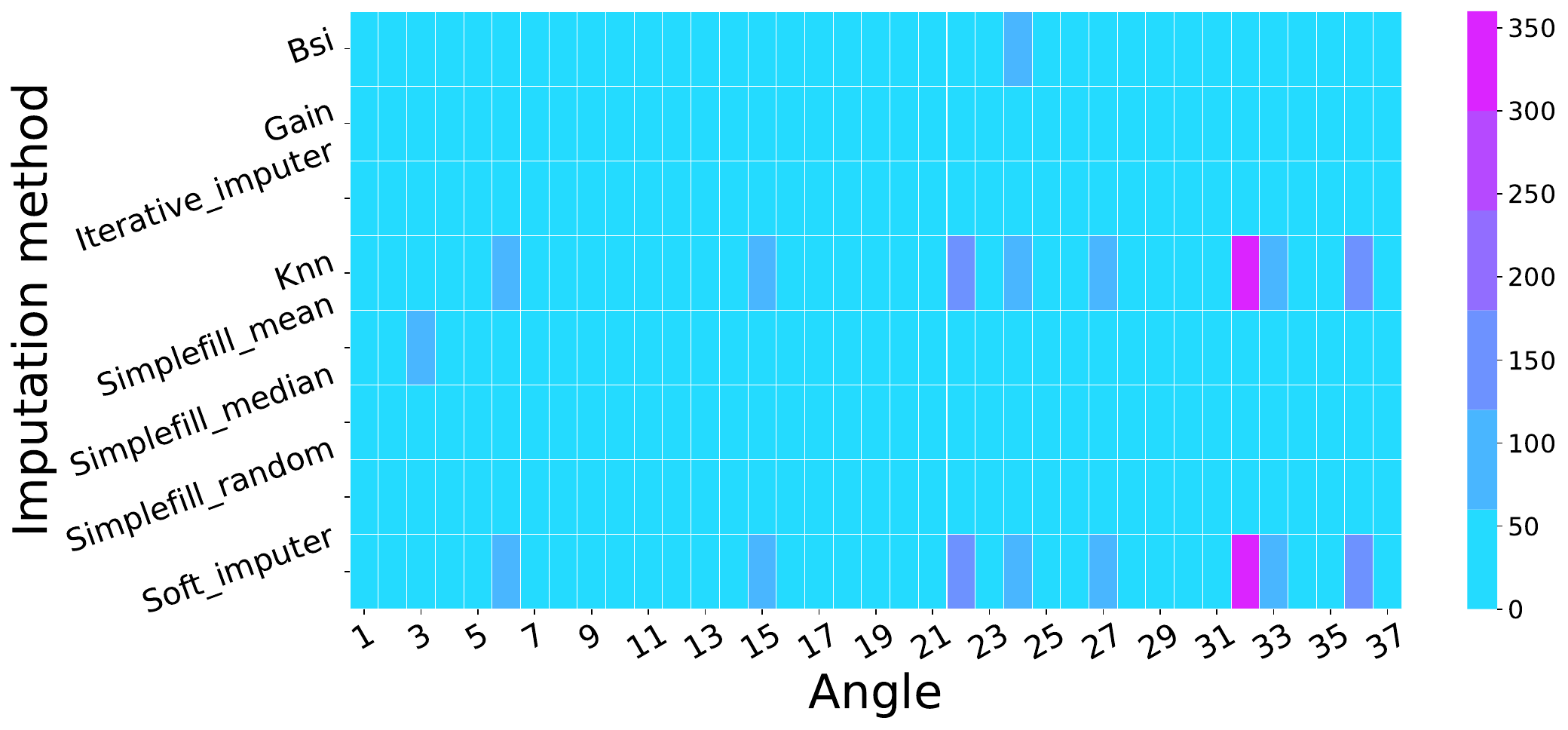}}
    \caption{MAE for univariate data imputation of single player for three datasets with. \label{fig:Single_Variate} }
\end{figure}

For randomly missing (MCAR) points, which were the simplest, there were some methods that were perform reasonably well. The mean absolute error in Fig. \ref{fig:Single_0} is confined to a relatively low range, with the color bar selection ranging from 0 to 70. in this case, this is expected as this scenario was both, simple and could be interpolated with the adjacent temporal data points.

However, the challenge intensifies significantly with more complex missingness. When the data is missing for critical transitions, Fig. \ref{fig:Single_1} shows the mean absolute errors increased significantly, with the largest value associated to the MAE exceeding thresholds upwards of 350. This scenario clearly differentiates simple statistical methods from the advanced models due to their critical interaction with time. The simple methods such as our SimpleFill Mean and SimpleFill Median perform poorly, as both of these methods their core mechanism—averaging— both methods flattens the critical peaks and valleys that defined the motion dynamics and create a incorrect dependency only on this time series.

The dataset that includes sequences of consecutive missing values located around transition points, showed the largest error values as shown in Fig. \ref{fig:Single_2}. This demonstrates, again, the difficult task of accurately imputing large portions of data. Observably, GAIN and Iterative Imputer, sophisticated models that learn non-linear relationships within the sequence, performed better since they could utilize their complex modeling capacity to more effectively approximate the underlying dynamics of the sequence, and therefore, significantly reduce their mean error relative to simpler methods.

In summary, as the missing data conditions get more complicated, the results of the error values across the datasets clearly show a gradient of increasingly more difficult and erroneous data as depicted in Fig. \ref{fig:Single_Variate}. The ability of imputation techniques to replicate and adjust to the fundamental dynamics of the motion sequences determines their performance mostly. When comparative external data was not available, more sophisticated techniques including GAIN and iterative Imputer were able to control the variance and complexity underlying intricate motion data. In summary, this study shows that although selecting simple imputation techniques based on the properties of the missing data may be crucial, more complicated approaches, such GAIN, have the possibility to improve data reconstruction integrity under challenging data conditions.

Table \ref{tab:missing_single_player} lists the MAE for various univariate imputation models employed on single-player motion capture data. All of the presented results demonstrated the inherent difficulty associated with reconstructing skill-based kinematic data where no other data is utilized. Both KNN and Soft Imputer exhibited the best accuracy in imputation when the amount of missing data was low (5\%), and would have likely leveraged the closest available reassembled temporal patterns elsewhere in the same sequence for reconstructing the data. 

These approaches performed well at lower levels of missing data, but inevitably the performance declined as the amount of missing data increased, indicating that losing larger sections of important dynamic information also compromises the temporal local context. Conversely, the most basic set of models,  statistical methods, such as Simplefill random has the least accurate rates in Table \ref{tab:missing_single_player}, no matter the percentage of missing data, which would be expected since this method does not consider movement data's basic biomechanical and temporal structure whatsoever. For the ML and DL methods, both GAIN and iterative machine learning approaches each exhibited competitive yet inconsistent performance across many cases, often trailing basic simpler procedures in accuracy at differing levels of information loss. This discussion depicts another important note made regarding single-player analysis. Without using the cross-sectional context of other context of other players, imputation reliability, or sensitivity, becomes highly dependent to the percentage of missing data, and no one method ever truly beat the task of reconstructing complex, and structured dynamic movements.

 \begin{figure}[H]
        \subfloat[Bsi (MAE:3.31)\label{fig:random_points_true_bsi}]{
            \includegraphics[width=0.5\linewidth]{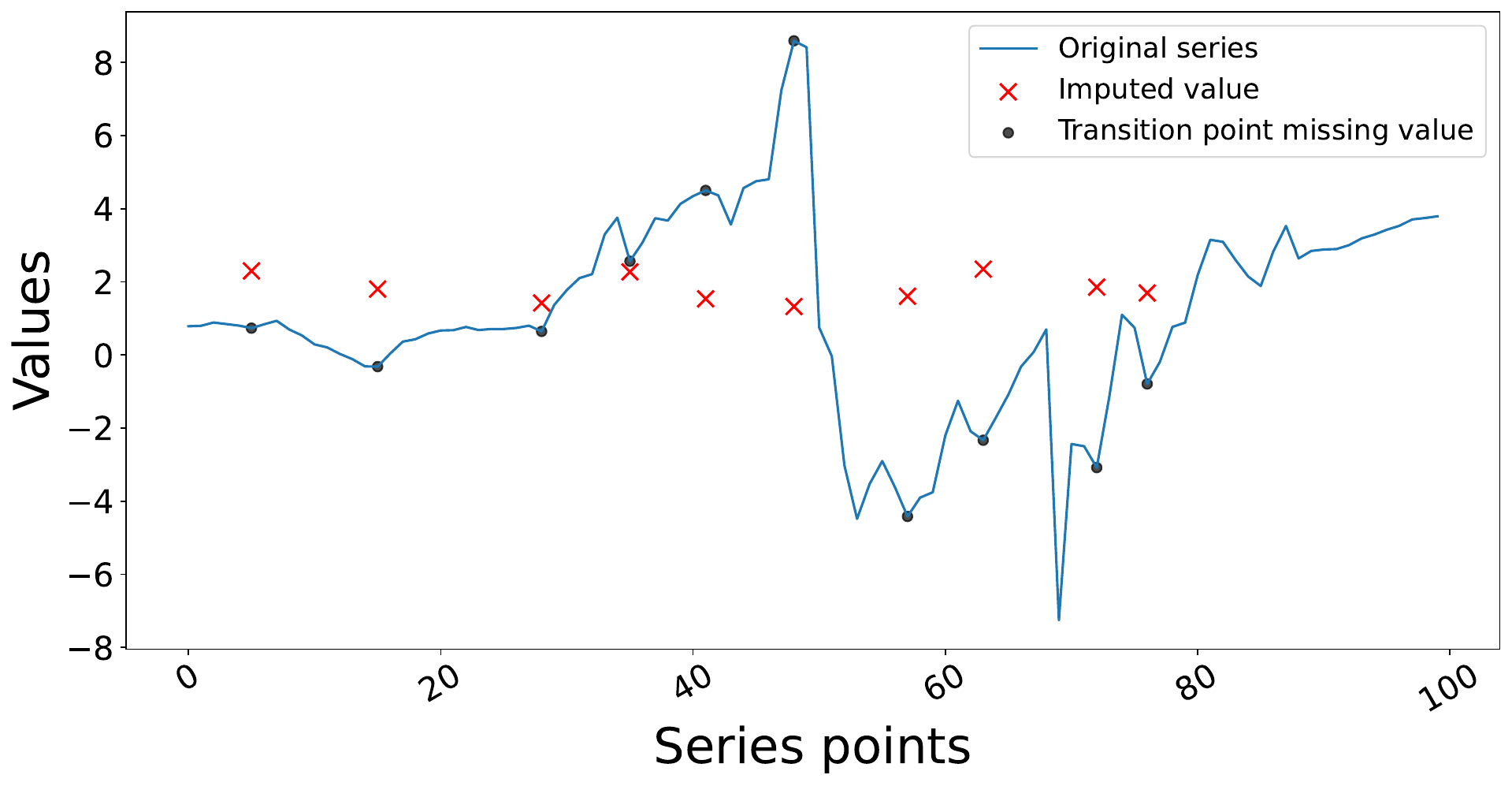}}
~
		\subfloat[GAIN (MAE:2.98)\label{fig:transition_points_true_gain}]{
            \includegraphics[width=0.5\linewidth]{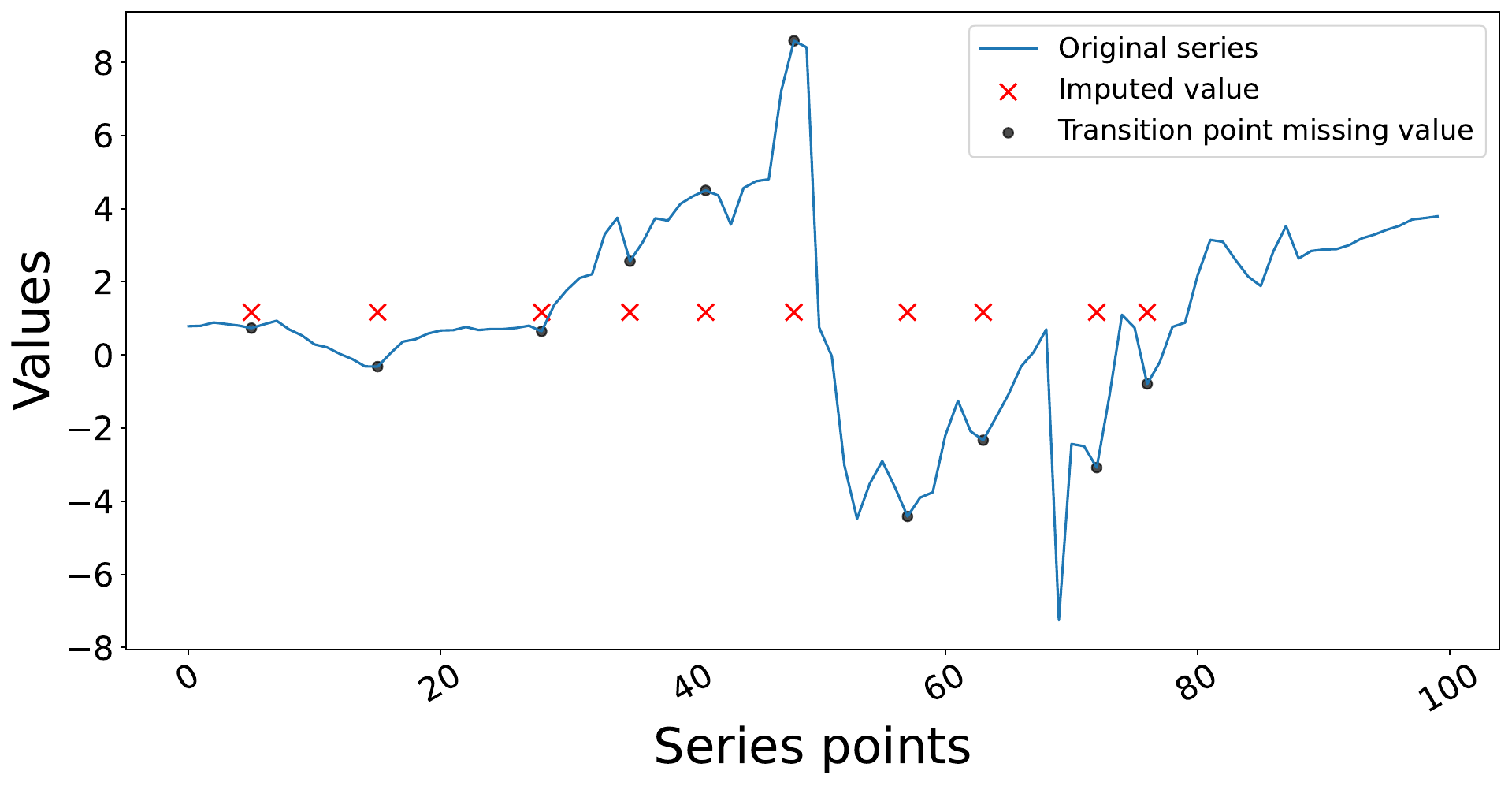}}

        \subfloat[Iterative imputer (MAE:2.98)\label{fig:random_points_true_itrimp}]{
            \includegraphics[width=0.5\linewidth]{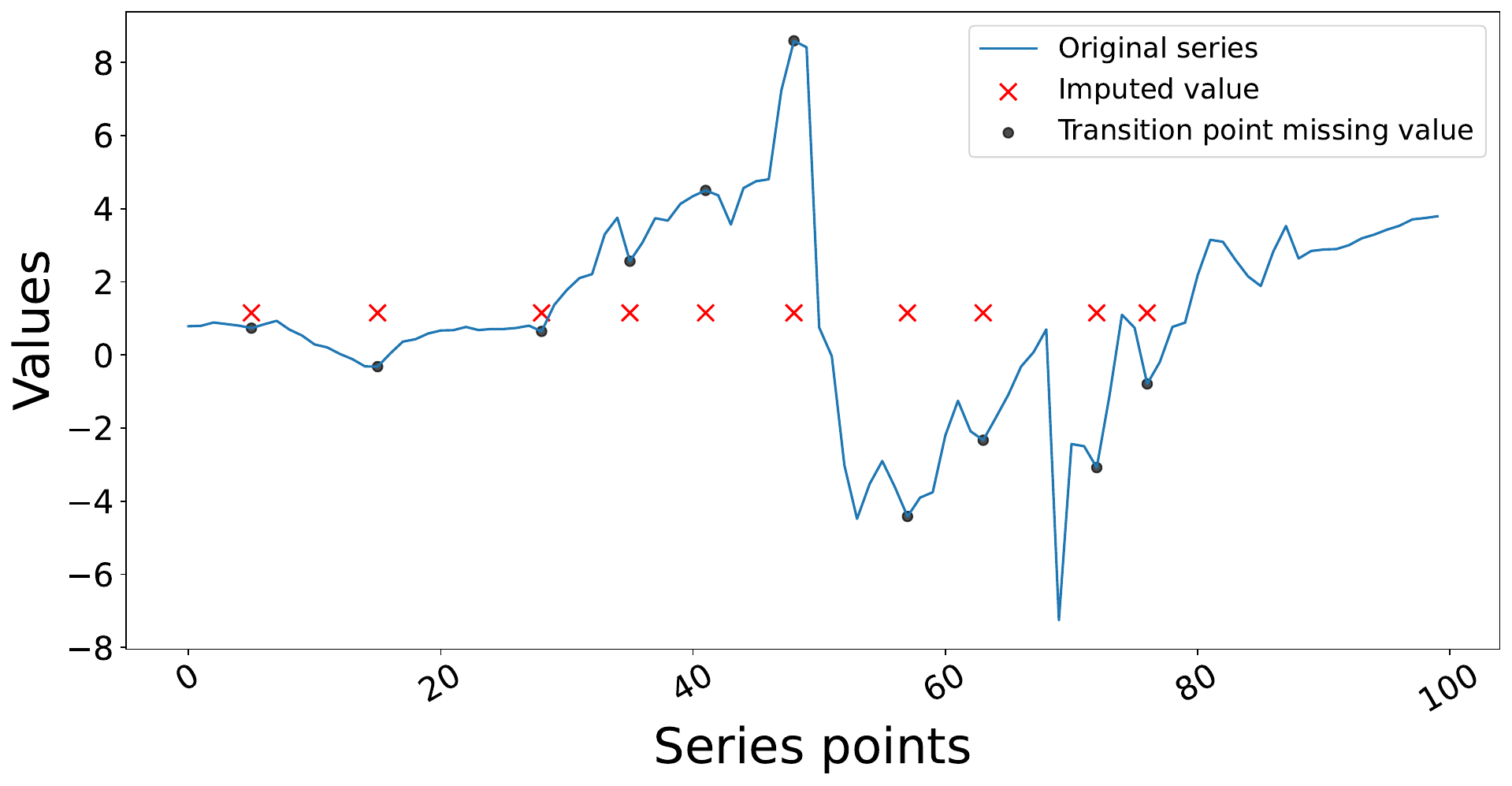}}
~
		\subfloat[KNN (MAE:2.79)\label{fig:transition_points_true_KNN}]{
            \includegraphics[width=0.5\linewidth]{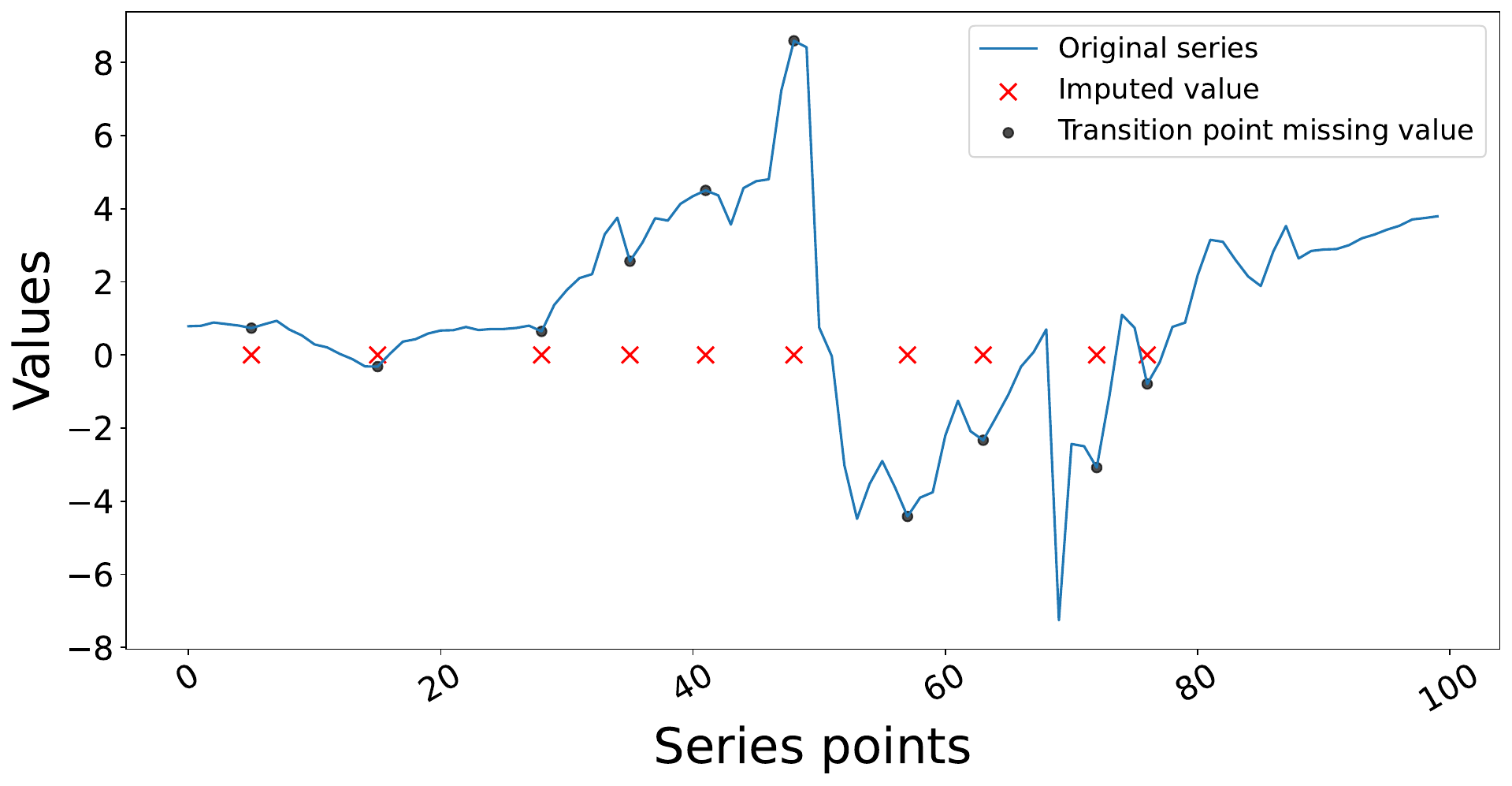}}

        \subfloat[Simplefill mean (MAE:2.15)\label{fig:random_points_true_smpfill_mean}]{
            \includegraphics[width=0.5\linewidth]{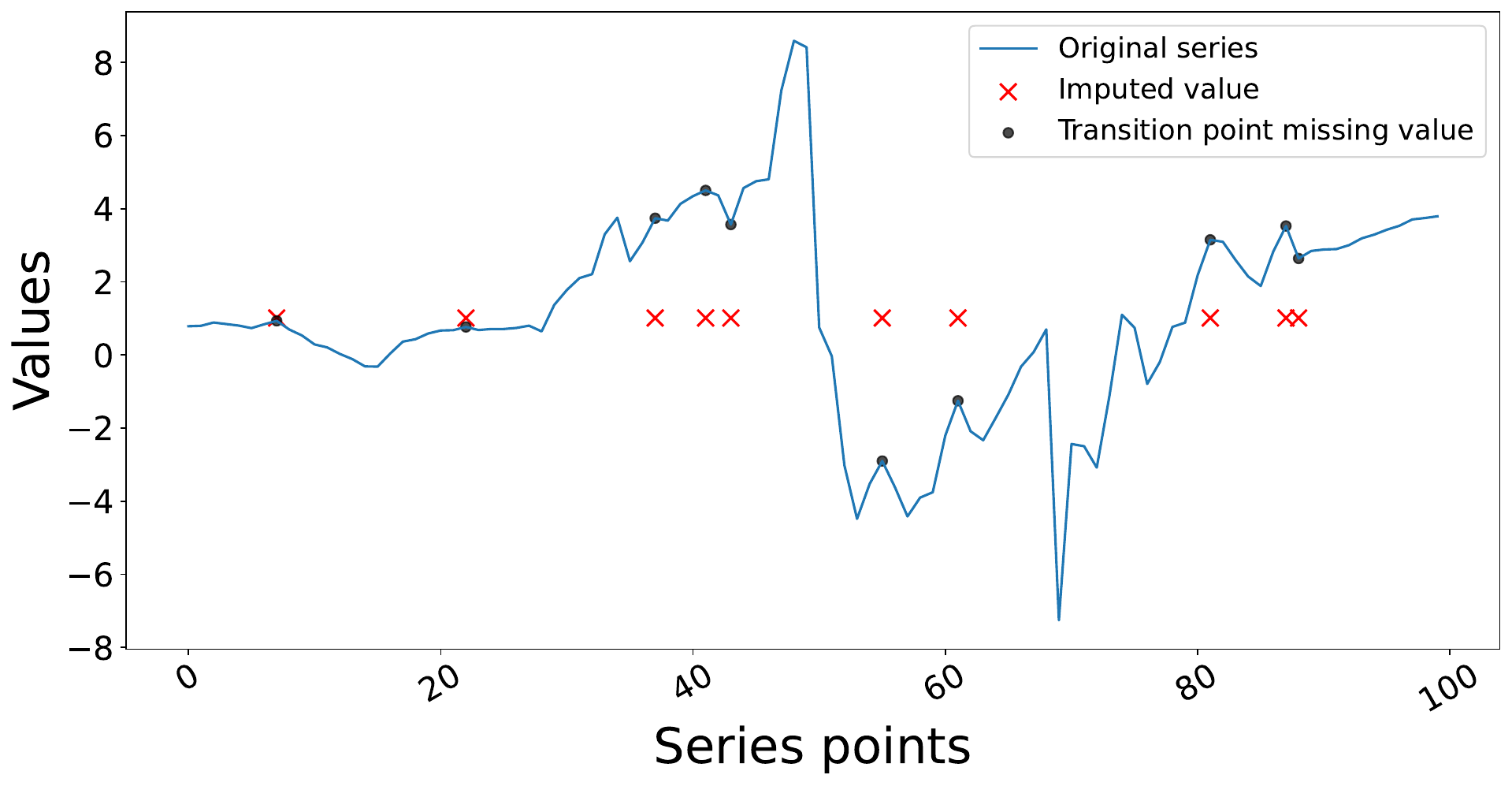}}
~
		\subfloat[Simplefill median (MAE:2.24)\label{fig:transition_points_true_median}]{
            \includegraphics[width=0.5\linewidth]{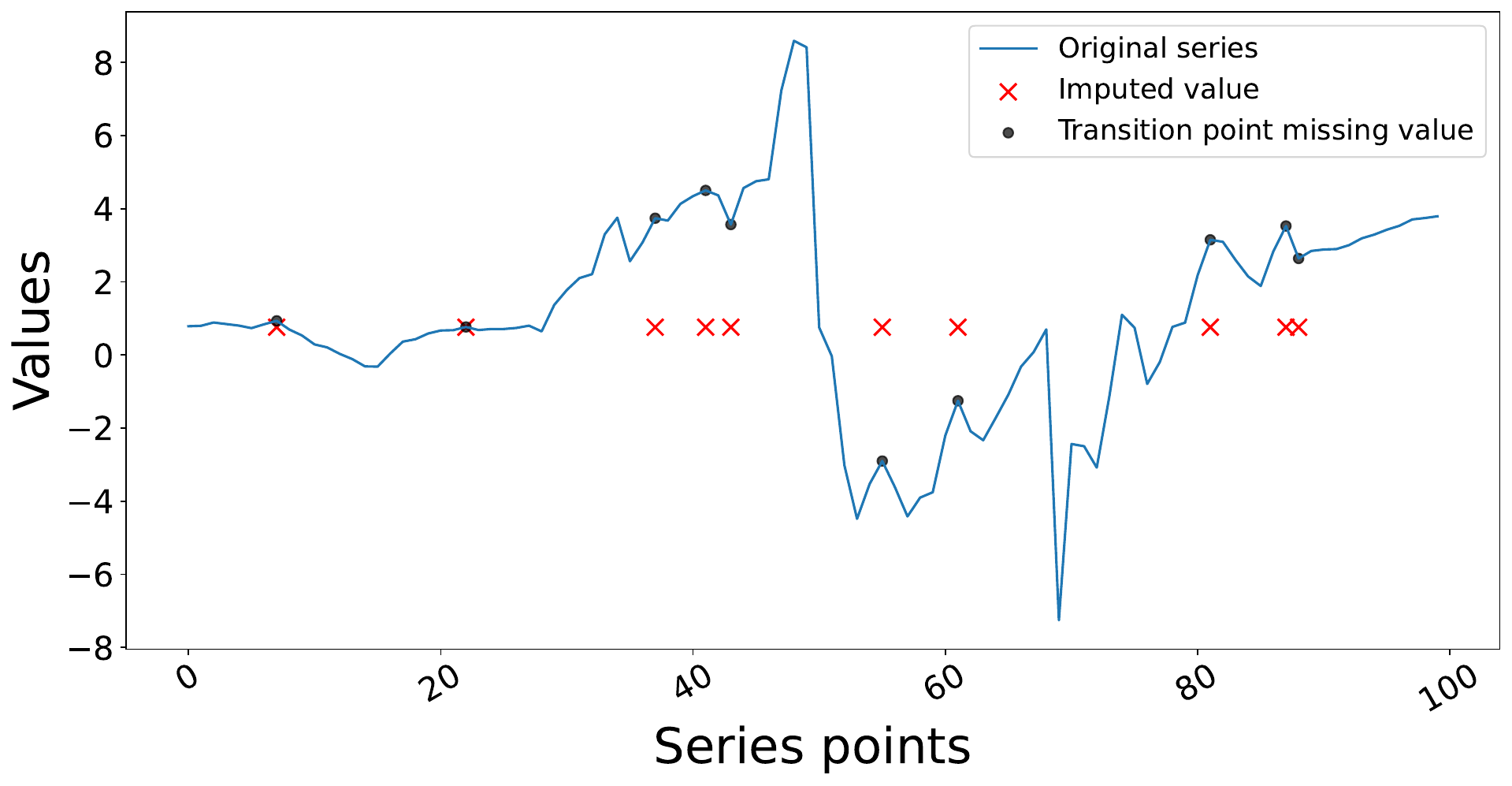}}

        \subfloat[Simplefill random (MAE:2.67)\label{fig:random_points_true_smpfill_rand}]{
            \includegraphics[width=0.5\linewidth]{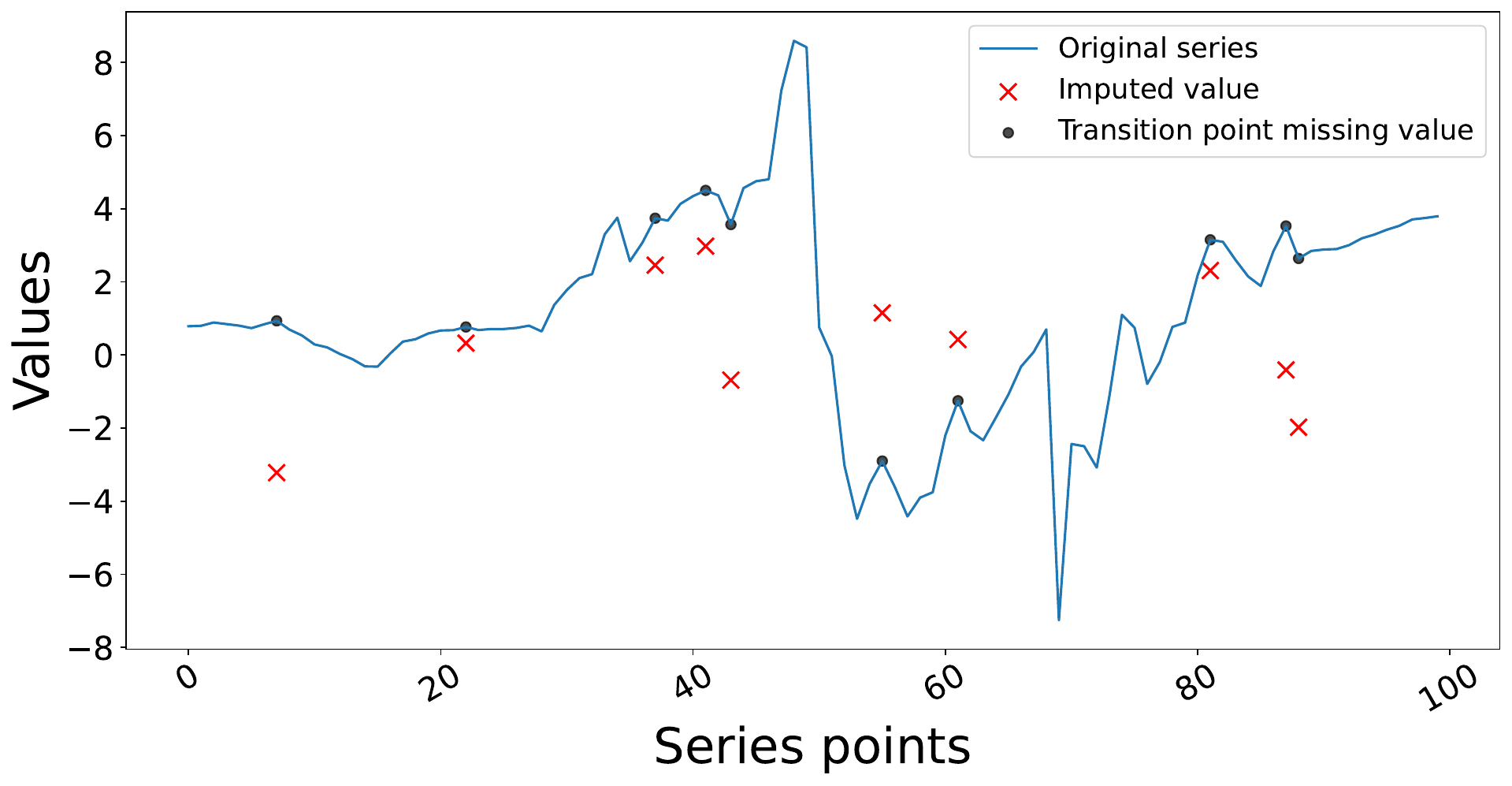}}
~
		\subfloat[Soft imputer (MAE:2.79)\label{fig:transition_points_true_soft_imp}]{
            \includegraphics[width=0.5\linewidth]{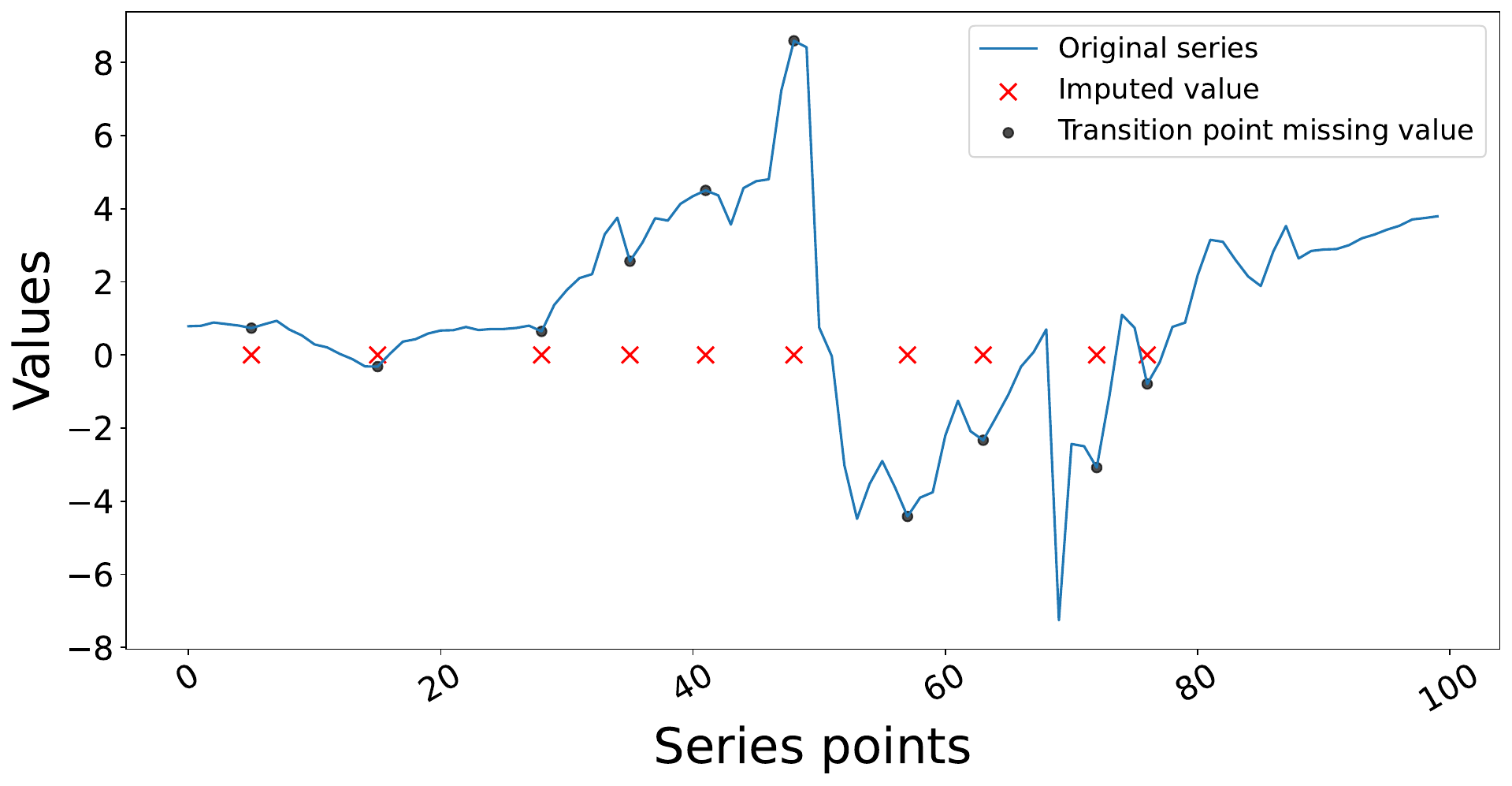}}            

    \caption{Imputed vs true missing values for univariate single-player series imputation (Transition missing points).
    % of a single data series. 
    \label{fig:Multiangle_true} }
\end{figure}  

% ----------------------- 
\begin{table}
\centering
\caption{MAE for univariate data imputation of Single-player for different missing-points percentages (Transition missing points).\label{tab:missing_single_player}}
\footnotesize
\begin{tabular}{|c|c|c|c|c|c|c|c|}
\hline
\multicolumn{1}{|l|}{\multirow{2}{*}{Imputation method}} & \multirow{2}{*}{Type} & \multicolumn{6}{c|}{Percentage of missing data} \\ \cline{3-8} 
\multicolumn{1}{|l|}{}                                   &                       & \multicolumn{1}{c|}{5\%}                                       & \multicolumn{1}{c|}{10\%}                                      & \multicolumn{1}{c|}{15\%}                                      & \multicolumn{1}{c|}{20\%}                                     & \multicolumn{1}{c|}{25\%}                                      & 30\%                                      \\ \hline
\hline
BSI & ML & 10.88 $\pm$ 5.99 & 11.12 $\pm$ 5.6 & 9.52 $\pm$ 5.12 & \textbf{9.69 $\pm$ 5.74} & 11.45 $\pm$ 5.74 & 10.0 $\pm$ 6.57 \\\hline
GAIN & DL & 11.99 $\pm$ 8.25 & 10.69 $\pm$ 6.0 & 13.51 $\pm$ 9.55 & 10.21 $\pm$ 6.73 & \textbf{10.8 $\pm$ 6.85} & \textbf{9.71 $\pm$ 7.2} \\\hline
Iterative imputer & ML & 11.93 $\pm$ 8.21 & 10.67 $\pm$ 6.06 & \textbf{9.5 $\pm$ 7.21} & 10.22 $\pm$ 6.82 & \textbf{10.8 $\pm$ 6.9} & \textbf{9.71 $\pm$ 7.19} \\\hline
KNN & ML & \textbf{8.73 $\pm$ 6.98} & 14.2 $\pm$ 10.09 & 9.56 $\pm$ 7.05 & 10.99 $\pm$ 8.8 & 12.0 $\pm$ 9.89 & 12.19 $\pm$ 8.92 \\\hline
Simplefill mean & Statistical & 11.93 $\pm$ 8.21 & 10.67 $\pm$ 6.06 & \textbf{9.5 $\pm$ 7.21} & 10.22 $\pm$ 6.82 & \textbf{10.8 $\pm$ 6.9} & \textbf{9.71 $\pm$ 7.19} \\\hline
Simplefill median & Statistical & 13.78 $\pm$ 9.8 & \textbf{10.05 $\pm$ 8.62} & 10.18 $\pm$ 9.33 & 10.75 $\pm$ 8.65 & 10.99 $\pm$ 8.57 & 9.97 $\pm$ 9.0 \\\hline
Simplefill random & Statistical & 20.79 $\pm$ 11.95 & 13.89 $\pm$ 8.18 & 18.74 $\pm$ 12.77 & 15.46 $\pm$ 12.66 & 17.24 $\pm$ 10.06 & 17.27 $\pm$ 12.73 \\\hline
Soft imputer & ML & \textbf{8.73 $\pm$ 6.98} & 14.2 $\pm$ 10.09 & 9.56 $\pm$ 7.05 & 10.99 $\pm$ 8.8 & 12.0 $\pm$ 9.89 & 12.19 $\pm$ 8.92 \\\hline

\end{tabular}
\end{table}

\subsection{Improving Imputation Accuracy with Multivariate Contexts}
Switching from a univariate to a multivariate context provides auxiliary information to the imputation models, allowing their performance to be greatly enhanced. Our approach distinguishes between two multivariate scenarios: information from data of different players performing the same skill (motion), and information from data of different angles from the same player.

\subsubsection{Across-Player Imputation: Leveraging Cohort-Based Correlations}

The Multivariate-Player context assumes that for a given kinematic variable (angle), the time-series from multiple players performing the same skill will share common patterns. This is a powerful assumption in sports science or clinical studies involving standardized tasks.

\begin{figure}[H]
        \subfloat[MAE for imputing randomly missing (MCAR) points.\label{fig:random_points_multi}]{
            \includegraphics[width=0.5\linewidth]{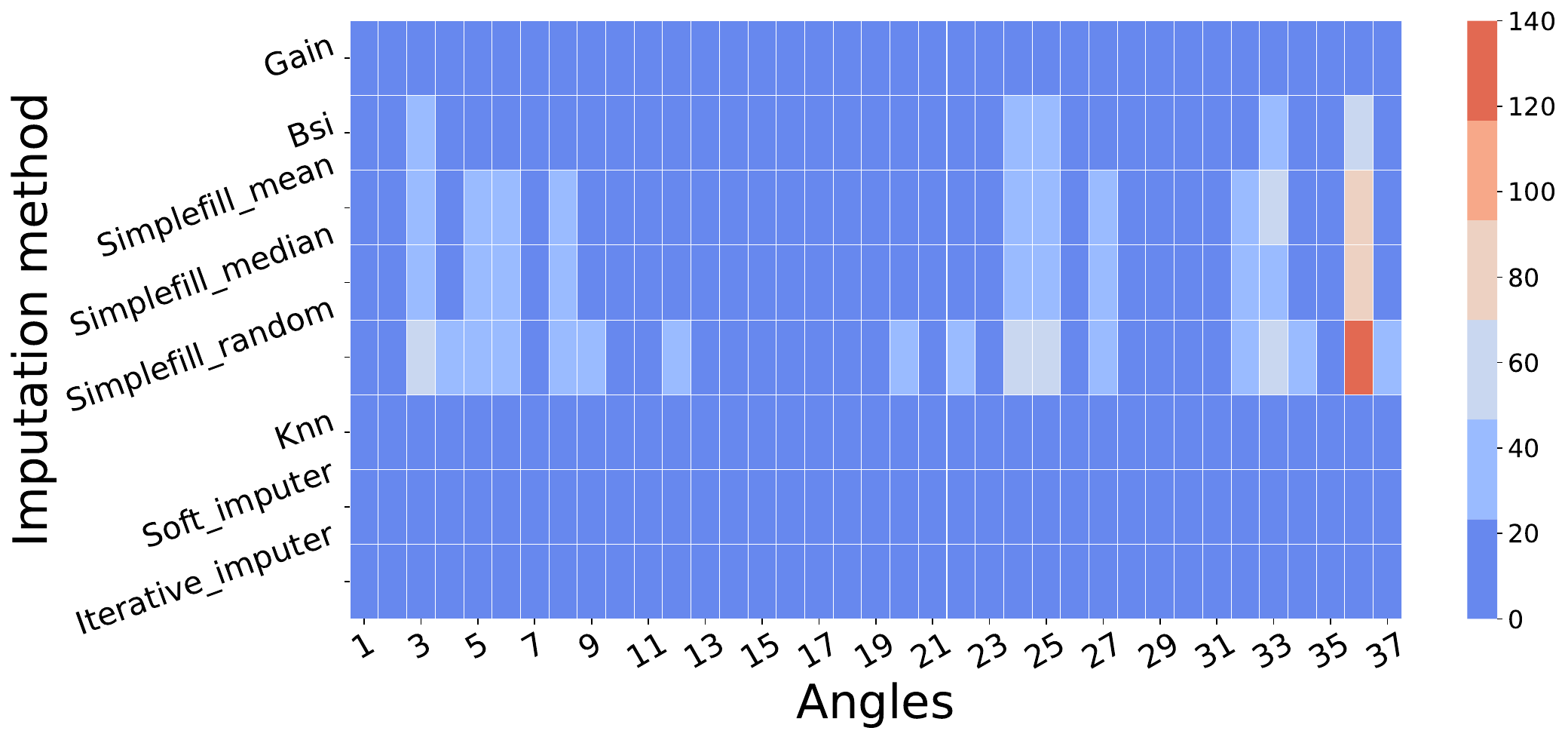}}
~
		\subfloat[MAE for imputing missing data of transition and random points.\label{fig:transition_points_multi}]{
            \includegraphics[width=0.5\linewidth]{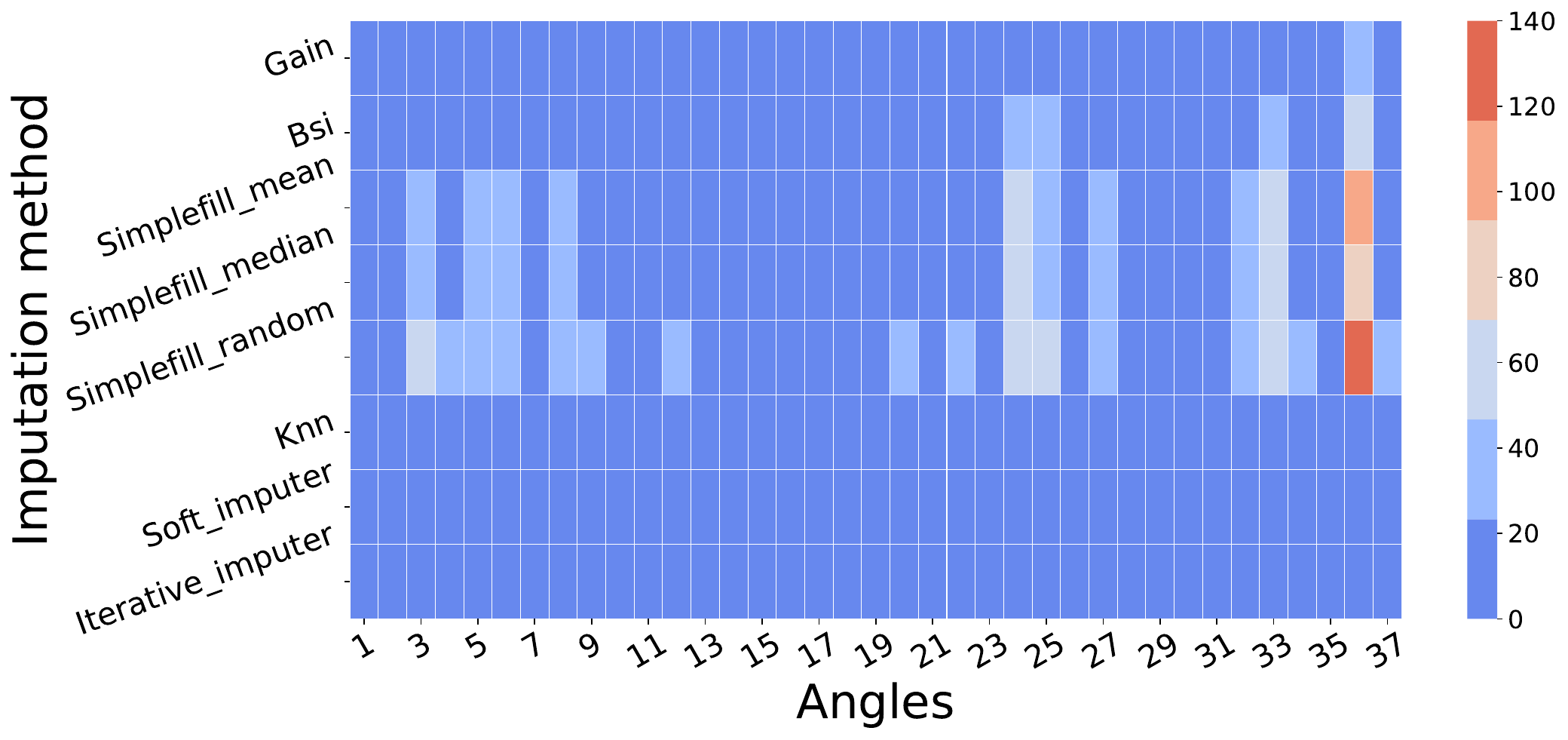}}

        \centering
        \subfloat[MAE for imputing sequences of consecutive missing data.\label{fig:interval_multi}]{
            \includegraphics[width=0.5\linewidth]{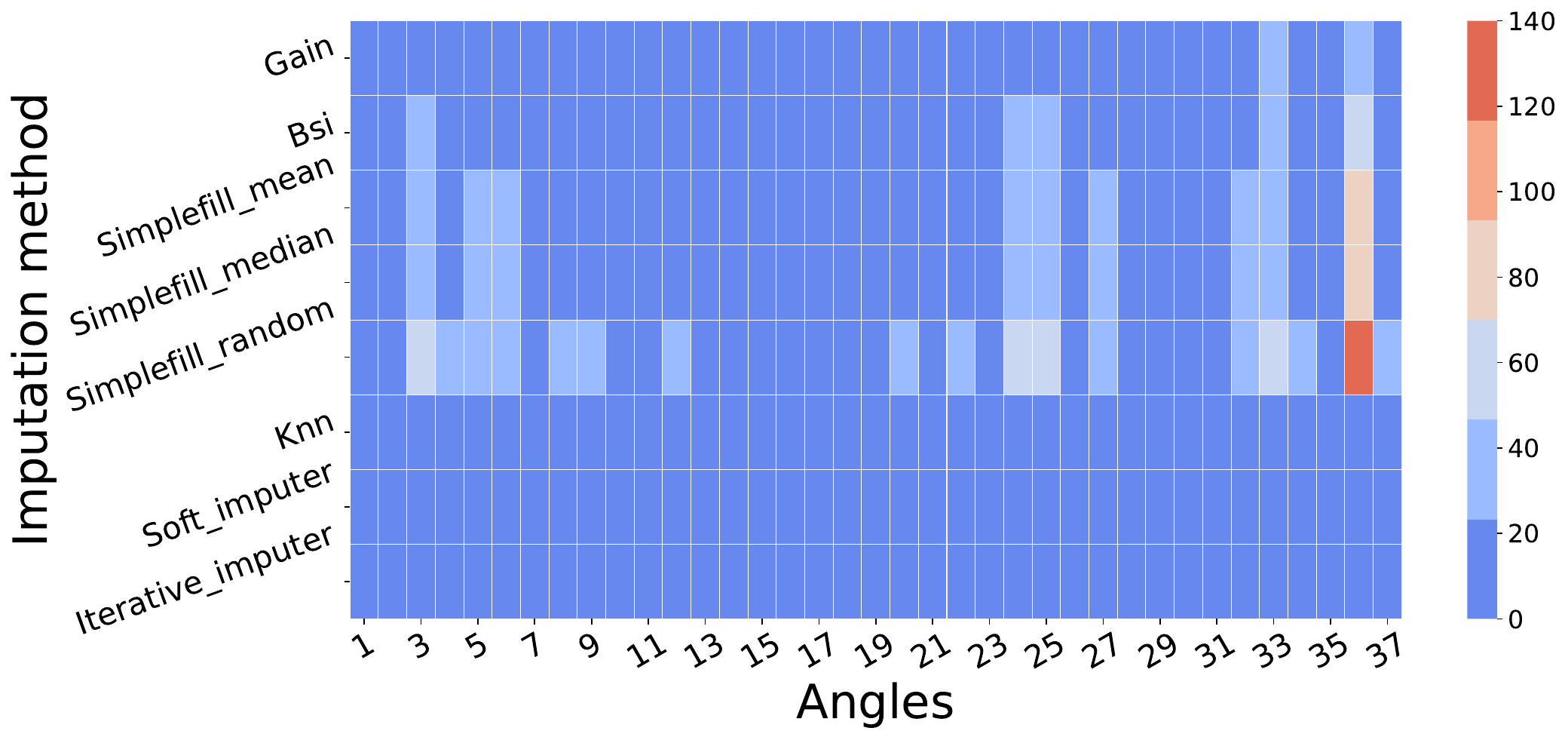}}
    \caption{MAE for multivariate data imputation using multi-player data for the same skill. \label{fig:Multi_player} }
\end{figure}

We investigated formulating the problem as a multivariate imputation task by utilizing similar sequences within the data, so possibly improving the accuracy and dependability of imputation. This method enables us to use relationships and patterns across several variables in every sequence, so providing a more informative context for imputing missing values. We sought to understand how different imputation techniques might take advantage of these interdependencies in scenarios spanning simple to highly complex missing data patterns by analyzing multivariate imputation across three datasets. In this experiment, we used the data of the same skill (motion) performed by several players to guide the imputation models to reconstruct the missing values for the players with missing motion data. The results of this experiment were depicted in Fig.   \ref{fig:Multi_player}.

The vast majority of error values for the first dataset, randomly missing (MCAR) points, are ranged from roughly 0 to 20 as shown in Fig. \ref{fig:random_points_multi}. While methods such as SimpleFill Random,   SimpleFill Mean, and SimpleFill Median preformed well in most cases, they struggled with only one case (i.e., angle 36) where the error values ranged from 60 to 140. With the multivariate context adding more information that helped even simpler techniques like SimpleFill Mean and SimpleFill Median, this range shows that most approaches performed effectively. More advanced techniques such KNN and Iterative Imputer used the correlations between variables to provide more exact imputations, so stressing the advantages of concurrently considering several variables.

With error values peaking at 140, the second dataset, when missing data is randomly distributed and at transition points,  results are depicted in Fig. \ref{fig:transition_points_multi}. The second dataset was even more complex and emphasized the need for multivariate techniques such as Soft Imputer and Iterative Imputer. Both advanced methods took advantage of the complicated interrelationships between several variables, and especially at transition points in the sequences where multiple variables are in motion, and these methods had to account for the multivariate complexity of the data to discretely estimate the missing values. In contrast, the SimpleFill Random technique had a harder time with the increasing complexity and was more consistent with the higher error rates associated with simple method approaches. Comparing Fig. \ref{fig:transition_points_multi} to Fig \ref{fig:Single_1}, we can notice that the error range decline from 350 to 140 that reflects the importance of the auxiliary information, other players' motion data, to the imputation models to improve error rates.

The third dataset of results shown in Fig. \ref{fig:interval_multi} presented substantial ranges of missing values even with transition points. Again, the error rate highest value reduced to 140 from 350, the highest value of the same dataset using the univariate imputation, Fig. \ref{fig:Single_2}.  In cases where advanced methods like GAIN and Iterative Imputer can use their unique modeling powers to accurately reconstruct these portions of missing data, we noted that they performed the best which is reassuring in that the extremely challenging continuity and accuracy of our imputed sequences relied on these methods management of the multivariate aspect of data.

Moving to a multivariate imputation framework significantly improves the imputation process, especially in complex scenarios. In a multivariate framework, multiple variables can be treated simultaneously which further leverages the data available with the benefit of varying dimensions of detections. There is an emphasis, however, in selecting a method based on the dimensionality of the data, and the complexities of missingness across multiple variables. Multivariate imputation techniques will yield notable advantages, being able to utilize and analyze the integrated data structures maintaining the structural and dynamic integrity of complex data or datasets.

The increments in accuracy are explained in Tables \ref{tab:missing_single_player} and \ref{tab:missing_multiplayers}. For example, with 25\% data missing at transition points, the MAE for the Iterative Imputer reduced from 10.8 ± 6.9 (Table \ref{tab:missing_single_player}, Single-Player) to an improved 5.82 ± 5.52 (Table \ref{tab:missing_multiplayers}, Multi-Player). The MAE for the KNN imputer reduced from 12.0 ± 9.89 to 4.29 ± 7.06, which is close to a 50 percent improvement. This proves that if while one player has a sensor that fails, the actions of other players provide a useful information to guide the imputation model for a better imputed values of the the missing data. 

\begin{figure}[H]
        \subfloat[BSI (MAE:1.26)\label{fig:random_points_mult_true_bsi}]{
            \includegraphics[width=0.5\linewidth]{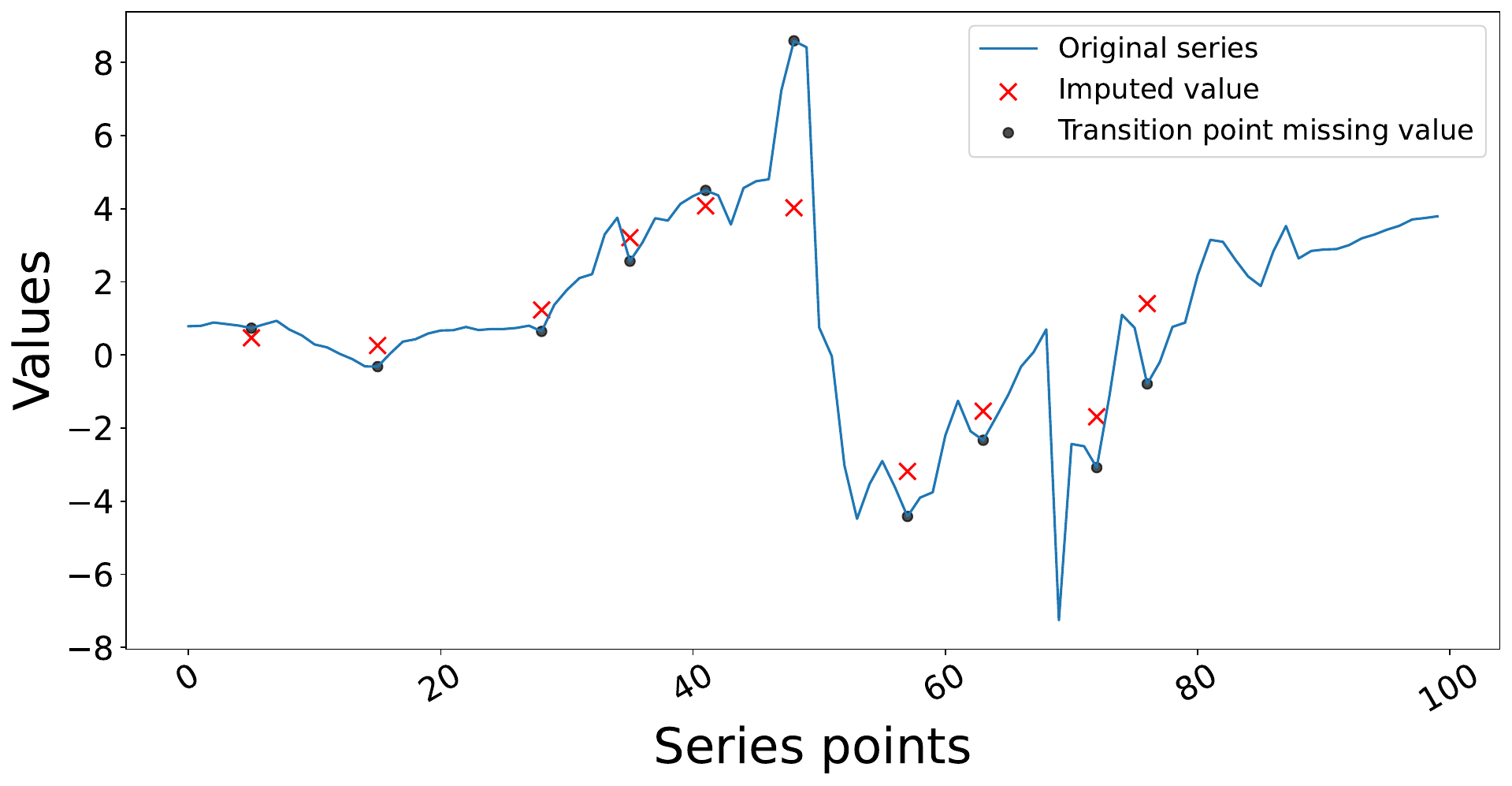}}
~
		\subfloat[GAIN (MAE:1.31)\label{fig:transition_points_mult_true_gain}]{
            \includegraphics[width=0.5\linewidth]{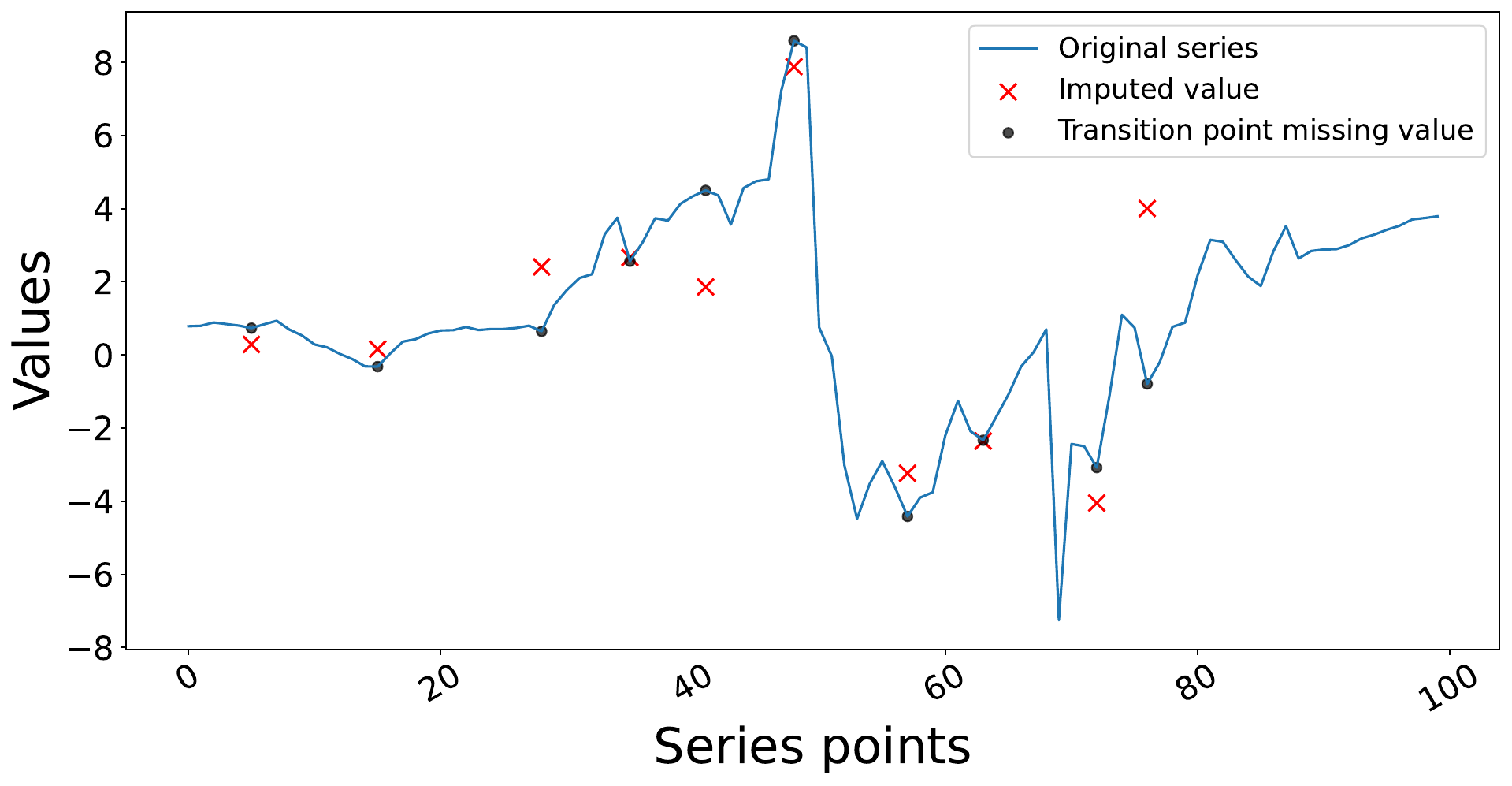}}

        \subfloat[Iterative imputer (MAE:1.19)\label{fig:random_points_mult_true_it_imp}]{
            \includegraphics[width=0.5\linewidth]{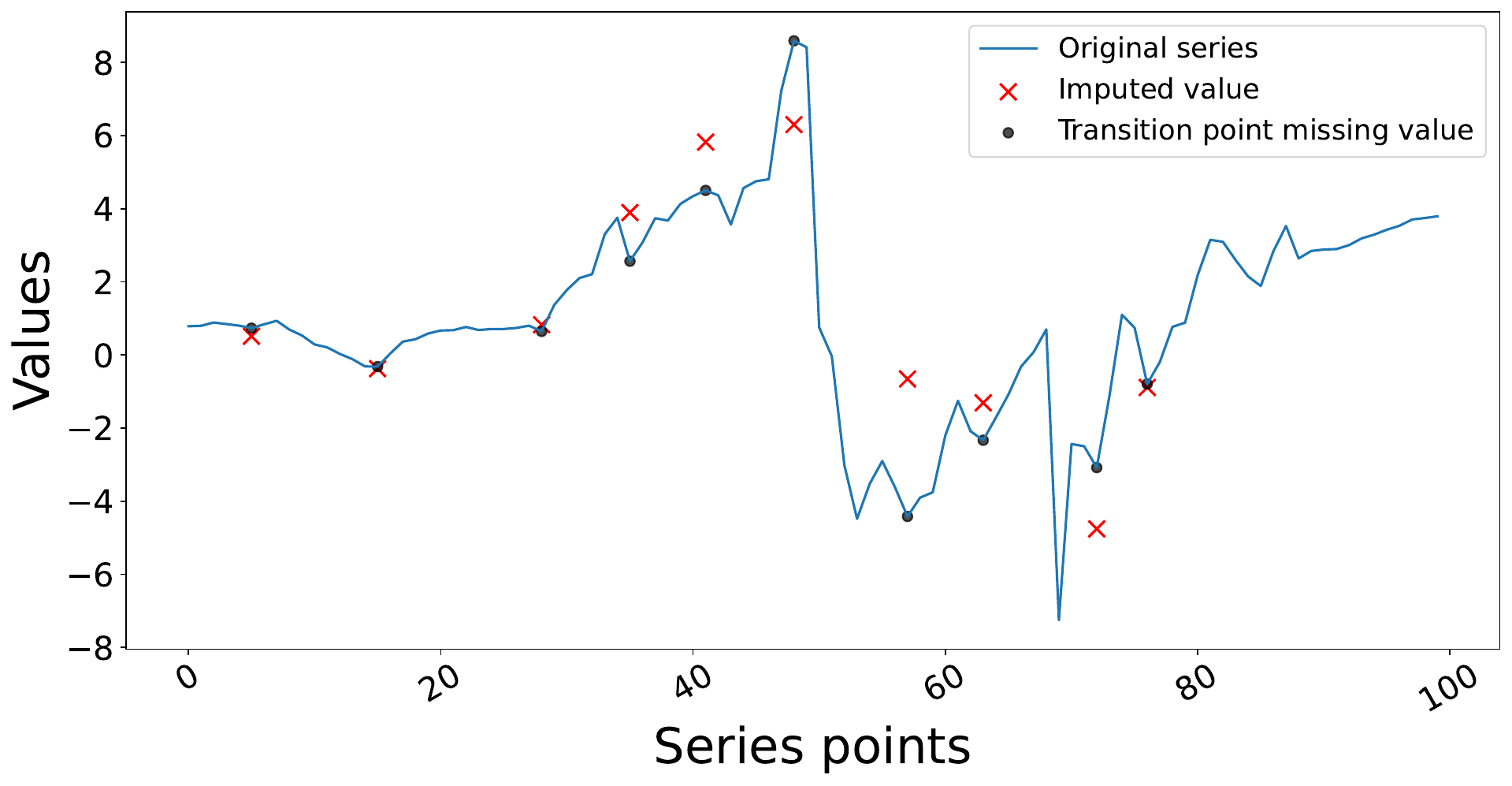}}
~
        \subfloat[KNN (MAE:1.07)\label{fig:random_points__mult_true_knn}]{
            \includegraphics[width=0.5\linewidth]{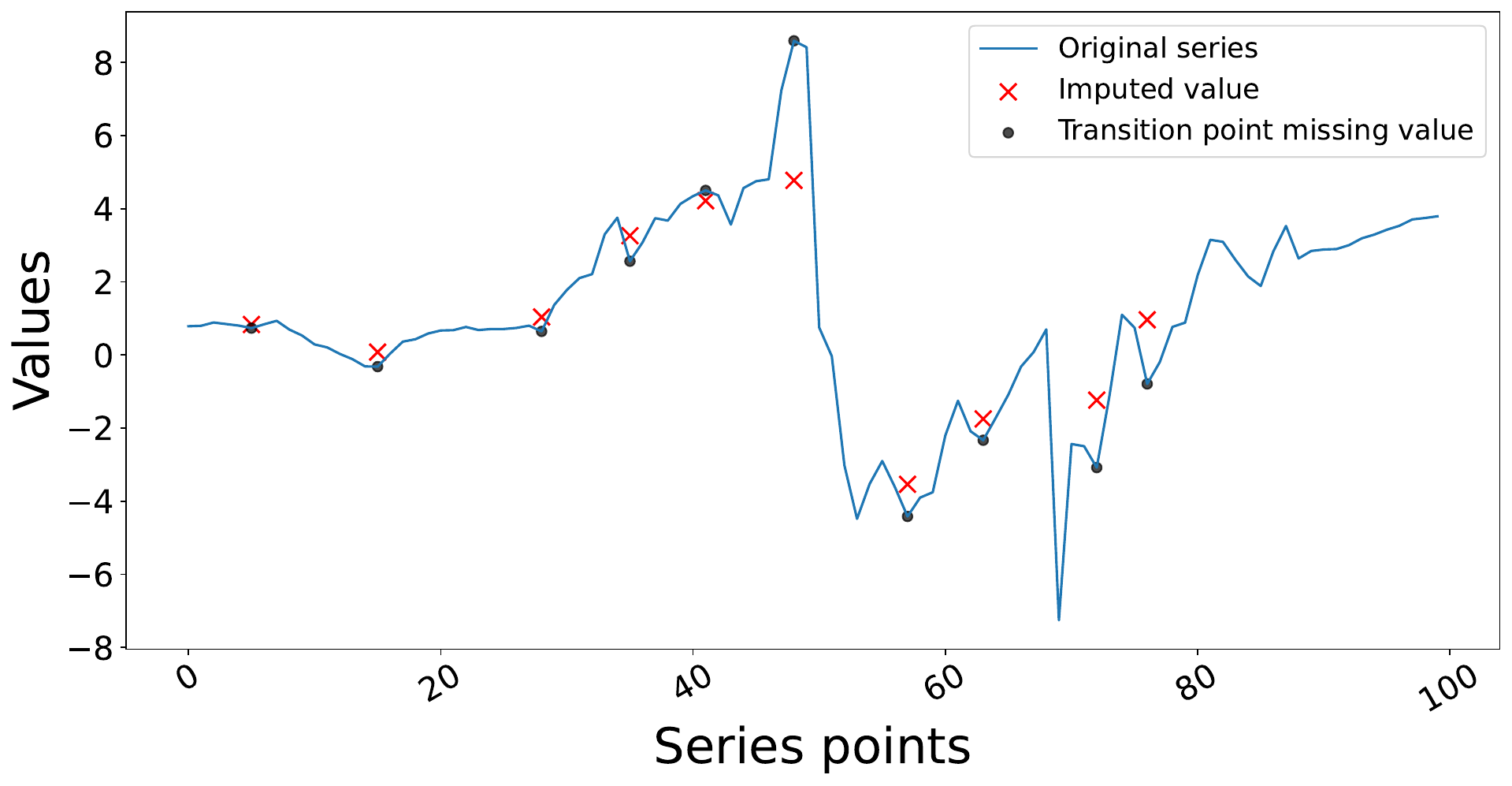}}
            
		\subfloat[Simplefill-mean (MAE:3.37)\label{fig:transition_points_mult_true_smp_mean}]{
            \includegraphics[width=0.5\linewidth]{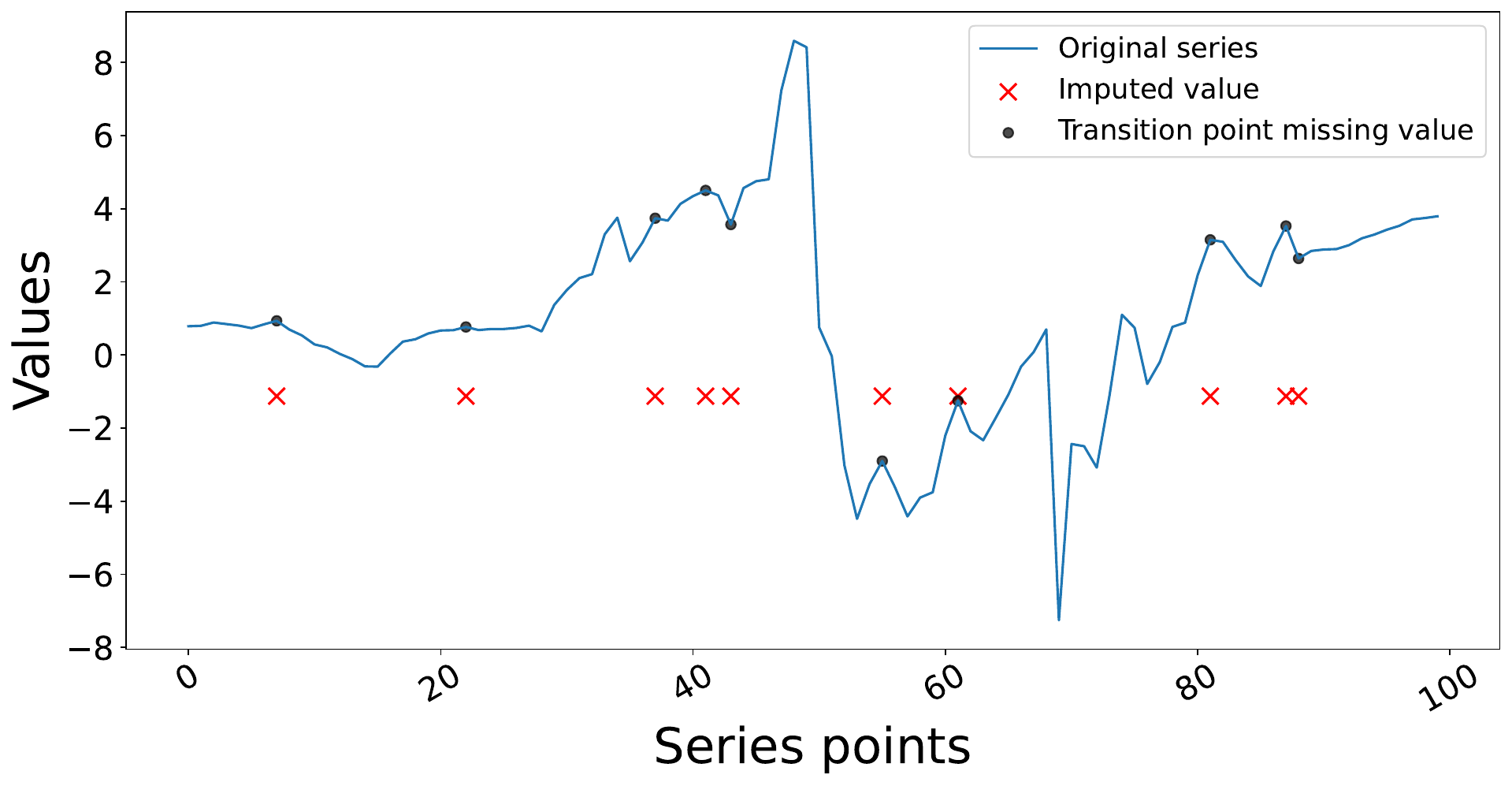}}
~
        \subfloat[Simplefill-median (MAE:3.34)\label{fig:random_points_mult_true_smp_median}]{
            \includegraphics[width=0.5\linewidth]{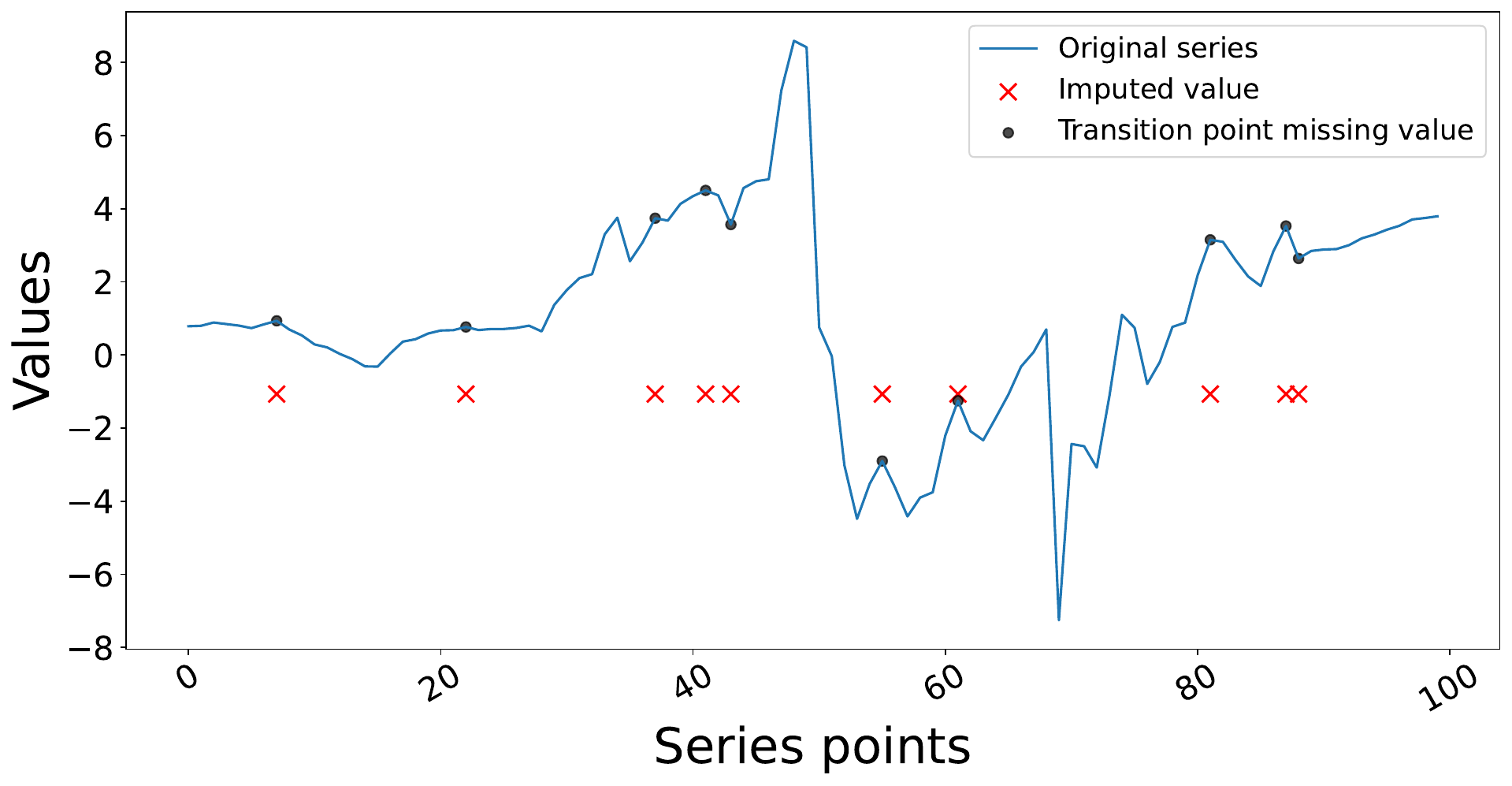}}

		\subfloat[Simplefill-random (MAE:3.70)\label{fig:transition_points_mult_true_smp_rand}]{
            \includegraphics[width=0.5\linewidth]{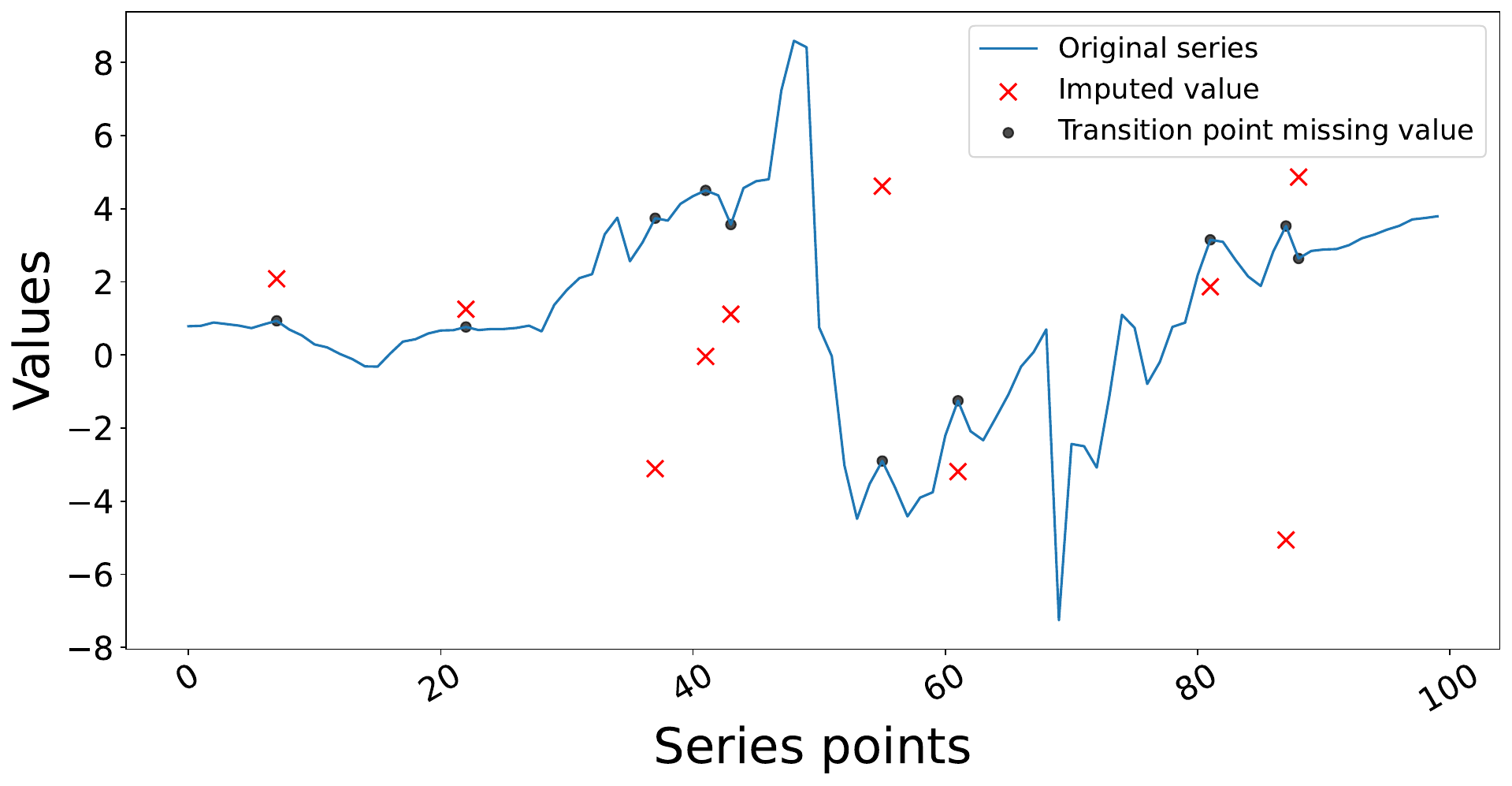}}     
~
        \centering
        \subfloat[Soft imputer (MAE:3.37)\label{fig:transition_points_mult_true_soft}]{
            \includegraphics[width=0.5\linewidth]{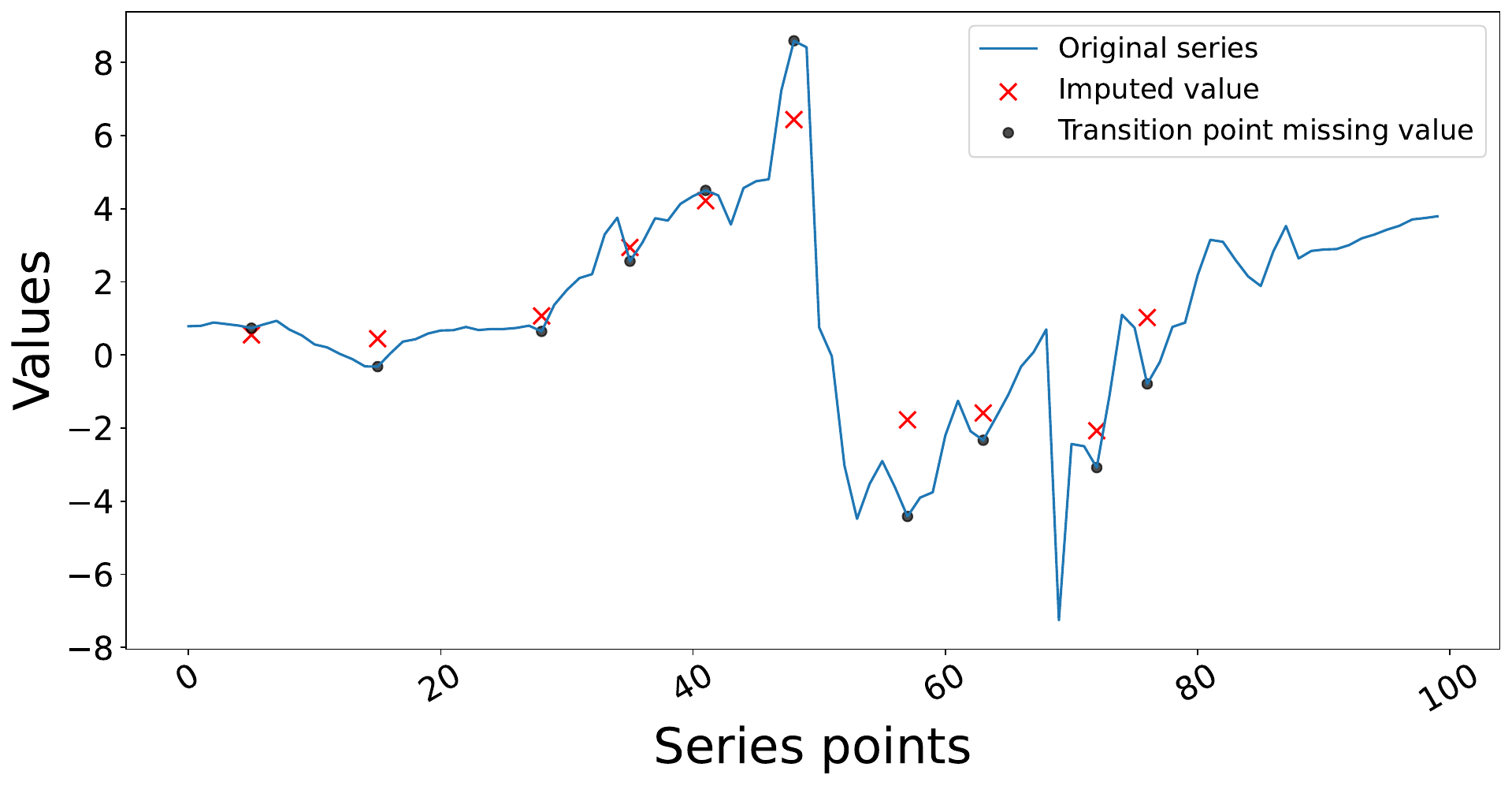}}          

    \caption{Imputed vs true missing values for multi-player imputation  (Transition missing points).
    \label{fig:Multivarite_multi_player} }
\end{figure}

A visual illustration of the imputed points and the true points for one case is depicted in Fig. \ref{fig:Multivarite_multi_player}. In Fig. \ref{fig:Multivarite_multi_player}, the imputed points by the statistical-based methods are far from the true points which indicates their weak performance, as shown in Figs. \ref{fig:transition_points_mult_true_smp_rand}, \ref{fig:random_points_mult_true_smp_median}, \ref{fig:transition_points_mult_true_smp_mean}. In contrast, Figs. \ref{fig:random_points_mult_true_bsi}, \ref{fig:transition_points_mult_true_gain}, and \ref{fig:random_points__mult_true_knn} show accurate data imputation, as the distances between the imputed points and the true points are very small.

% ----------------------- 
\begin{table}[]
\centering
\caption{MAE for multivariate data imputation of multi-player for different missing-points percentages (Transition missing points).\label{tab:missing_multiplayers}}
\footnotesize
\begin{tabular}{|c|c|c|c|c|c|c|c|}
\hline
\multicolumn{1}{|l|}{\multirow{2}{*}{Imputation method}} & \multirow{2}{*}{Type} & \multicolumn{6}{c|}{Percentage of missing data} \\ \cline{3-8} 
\multicolumn{1}{|l|}{} & & \multicolumn{1}{c|}{5\%} & \multicolumn{1}{c|}{10\%} & \multicolumn{1}{c|}{15\%} & \multicolumn{1}{c|}{20\%} & \multicolumn{1}{c|}{25\%} & 30\% \\ \hline\hline

BSI & ML & 6.47 $\pm$ 5.16 & 6.65 $\pm$ 5.47 & 7.21 $\pm$ 4.95 & 7.68 $\pm$ 4.99 & 8.48 $\pm$ 5.15 & 8.12 $\pm$ 5.21\\\hline
GAIN & DL & 2.21 $\pm$ 1.36 & 6.99 $\pm$ 7.67 & 2.89 $\pm$ 2.18 & 14.72 $\pm$ 6.89 & 6.2 $\pm$ 6.87 & \textbf{4.38 $\pm$ 5.45}\\\hline
Iterative imputer & ML & 0.86 $\pm$ 0.88 & 3.29 $\pm$ 4.13 & 2.74 $\pm$ 2.5 & 2.49 $\pm$ 2.42 & 5.82 $\pm$ 5.52 & 4.4 $\pm$ 4.33\\\hline
KNN & ML & \textbf{0.77 $\pm$ 0.87} & \textbf{2.19 $\pm$ 3.37} & \textbf{2.26 $\pm$ 3.46} & \textbf{2.48 $\pm$ 3.5} & \textbf{4.29 $\pm$ 7.06} & 6.96 $\pm$ 8.57\\\hline
Simplefill mean & Statistical & 10.96 $\pm$ 5.99 & 11.1 $\pm$ 4.97 & 9.08 $\pm$ 4.77 & 10.06 $\pm$ 4.72 & 10.8 $\pm$ 5.15 & 10.07 $\pm$ 4.76\\\hline
Simplefill median & Statistical & 10.3 $\pm$ 2.94 & 11.81 $\pm$ 5.13 & 8.92 $\pm$ 3.14 & 10.06 $\pm$ 3.97 & 10.89 $\pm$ 4.8 & 10.53 $\pm$ 4.21\\\hline
Simplefill random & Statistical & 9.12 $\pm$ 6.31 & 12.95 $\pm$ 8.19 & 18.75 $\pm$ 12.39 & 16.89 $\pm$ 12.8 & 15.27 $\pm$ 12.28 & 14.71 $\pm$ 13.44\\\hline
Soft imputer & ML & 3.06 $\pm$ 1.75 & 5.7 $\pm$ 7.89 & 2.99 $\pm$ 3.23 & 3.91 $\pm$ 5.24 & 5.87 $\pm$ 7.82 & 4.81 $\pm$ 6.63\\\hline

\end{tabular}
\end{table}

% ------------------------ True vs. imputed values multiplayer imputation

\subsubsection{Across-Angle Imputation: Exploiting Biomechanical Correlations}

The Multivariate-Angle context was built upon the principle of biomechanical coupling: for one subject (person), the movement of one joint is dependent on the movements of other joints. This was the most impactful context, especially with structured data loss.

\begin{figure}[H]
        \subfloat[MAE for imputing randomly missing (MCAR) points.\label{fig:multi_angles_random_points}]{
            \includegraphics[width=0.55\linewidth]{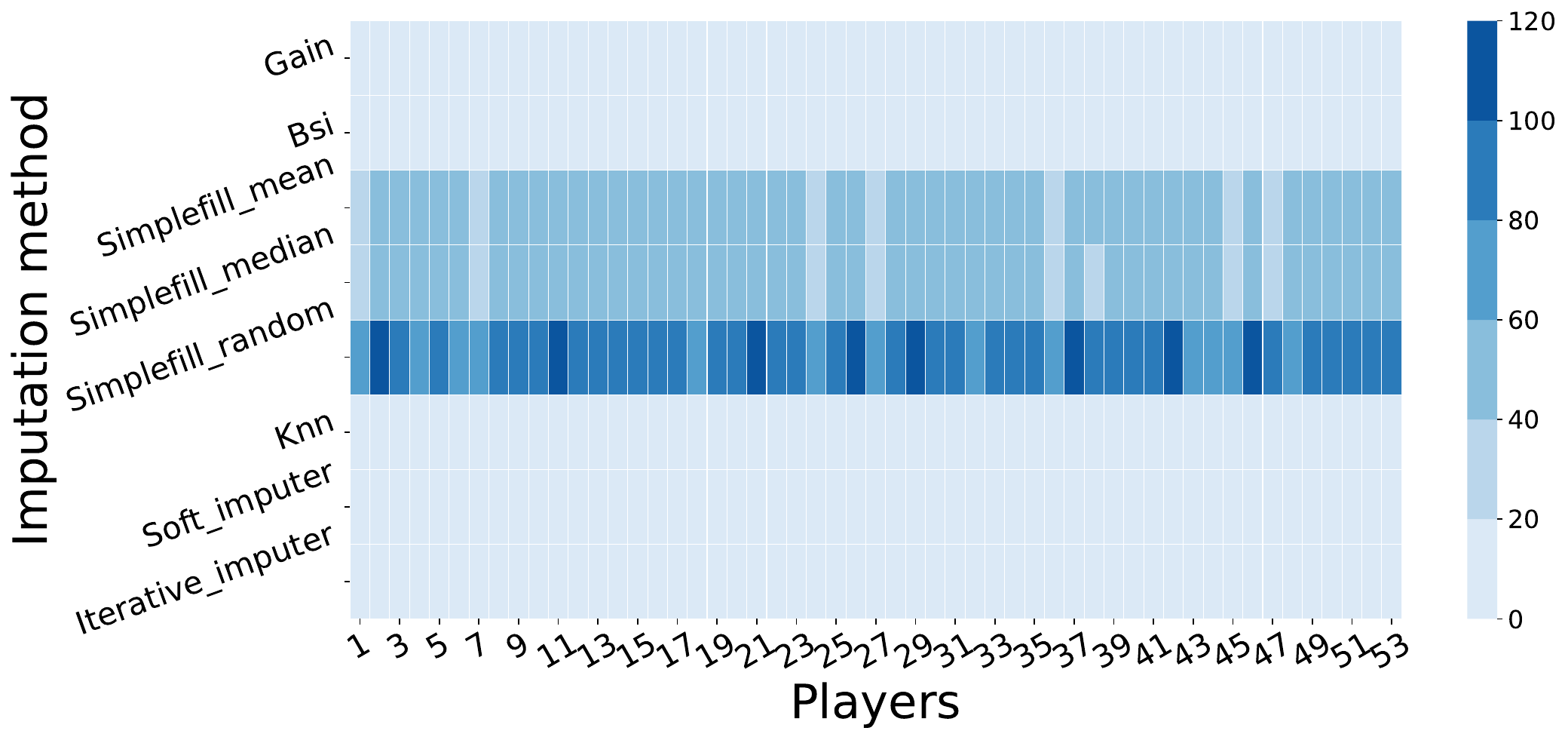}}
~
		\subfloat[MAE for imputing missing data of transition and random points.\label{fig:multi_angles_transition_points}]{
            \includegraphics[width=0.55\linewidth]{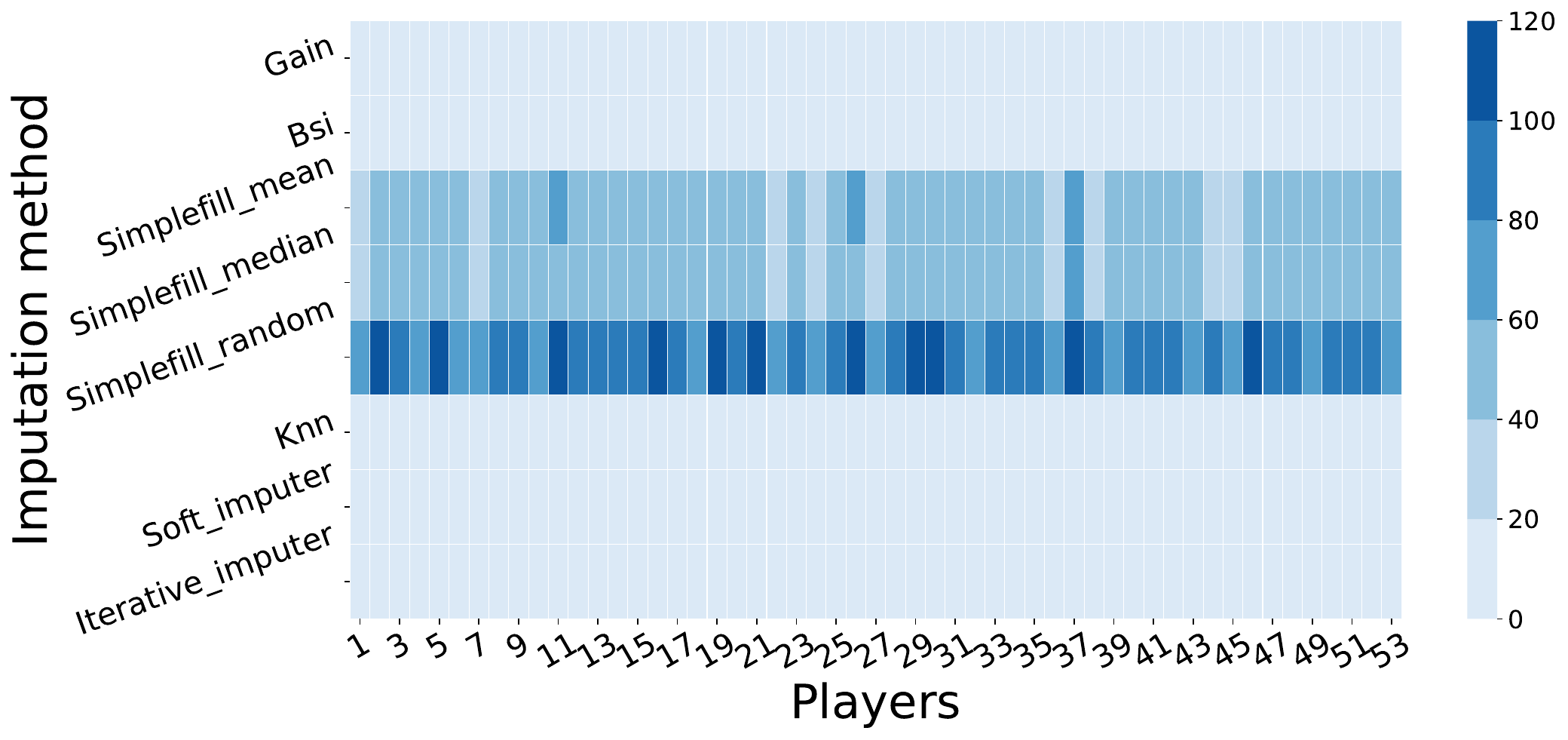}}

        \centering
        \subfloat[MAE for imputing sequences of consecutive missing data. \label{fig:multi_angles_interval}]{
            \includegraphics[width=0.55\linewidth]{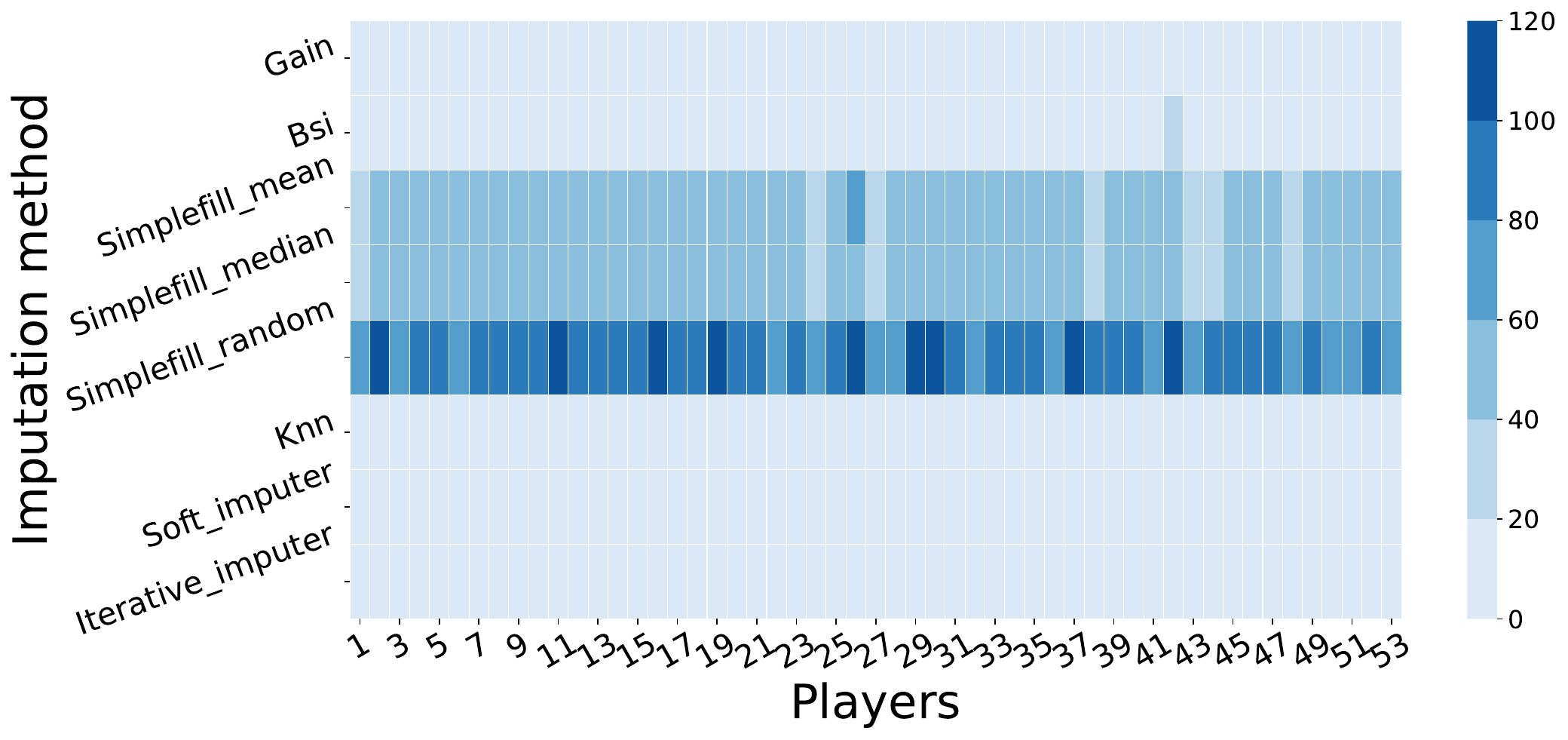}}
    \caption{MAE for multivariate data imputation using multi-angle data for the same player. \label{fig:Multiangle} }
\end{figure}

In the first dataset, sequences contained random missing values, and applying a univariate method resulted in an error range of 0 to 120, as depicted in Fig. \ref{fig:multi_angles_random_points}. Although the multivariate approach increased the complexity of the imputation procedure, the performance of the most of the non-statistical-based model greatly improved relate to the univariate results shown in Fig. \ref{fig:Single_0}. On contrary, the statistical based models performance declined. This behavior of model performance continues to appear on the other two datasets of missing values, the missing values around the transition points as in Fig. \ref{fig:multi_angles_transition_points} and the interval of missing values as in Fig. \ref{fig:multi_angles_interval}.

The second dataset contained random discrete missing values and missing values at transition points in the sequences, which was a more complex scenario, where univariate missing value imputation methods yielded error values of 0 - 350. The use of multivariate imputation methods drastically improved the error values to 0 - 120, as shown in Fig. \ref{fig:multi_angles_transition_points} . This represents more than 50\% reduction in max error and supports the effectiveness of multivariate methods in utilizing relationships among multiple variables when confronted with complex patterns of missing data, especially at critical transition points in sequences.

The third dataset exhibited the most significant difficulties, with long intervals of missing values including transition points. Although the multivariate approach still produced high error values, peaking at 120, as shown in Fig. \ref{fig:multi_angles_interval}. It was much better than the univariate results, so illustrating the advanced powers of techniques such as GAIN and Iterative Imputer in a multivariate setting.

Generally, especially in complicated situations, the shift to a multivariate framework shows a trend toward better imputation accuracy. Particularly where multivariate techniques can use inter-variable relationships to negotiate the complexity of missing data, thereby lowering error margins, the thorough comparison highlights great advantages. In disciplines like sports science, where exact data reconstruction is essential to the quality of performance analysis and consequent insights, this method is especially helpful.

\subsection{Discussion}
Our empirical investigation has first identified severe limitations of treating MoCap time-series as either merely sequences of observations or as univariate series. In this context, the imputation performance is critically dependent on the missingness mechanism, as for example; some simple, randomly missing points may reasonably interpolate from each side of the temporal sequence, however for more complex structures performance degrades severely. The failure of simple statistics methods on missing transitional points, is particularly telling and reinforces the nature of methods like mean-filling, which flatten some of the important peaks and valleys in the dynamics of the movement, thereby creating possibly unacceptable distortions. The extent to which this limitation is compounded in terms of instability for representative motions with block missingness, where there were no local conditions to view the interpolulated motion, forced the model to extrapolate without guidance and therefore resulting in fundamentally unreliable imputations	leading to fundamentally unreliable imputations.

The multi-variate frameworks offered a way to mitigate this issue by using critical auxiliary information, with the Across-Player context performing well also in incident related data loss. This context leverages cohort similarities and is predicated on assumptions that the different players would share common motions with certain skilled aspects. The model effectively learns a statistically shared template or archetype for the skill movement itself, by drawing information from a distribution of performances. The key performance observation is when a players sensor failed at a key incident in the movement (e.g. peak, valley) the equivalent motion from other players provided an excellent, high-fidelity basis for reconstructing the movement, explaining the significant error reduction for missing transition points.

Based on the varieties of the missing measurements, the Multivariate-Angle context consistently achieved the best performance, simply because it is the model most suited to the nature of MoCap data. The framework's ability to model the intrinsic biomechanical coupling of the human kinematic chain, and in that context it was very effective. For instance, based on the state of one joint (e.g., elbow) should have one state based on the states of the adjacent joints (e.g., shoulder and wrist). This context is particularly powerful when considering contiguous block missingness as well, it is because it is true that while temporal information is lost for one channel, the cross-sectional kinematic relationships with all other channels are intact for every time step. In short, this offers an ongoing and structural basis for reconstructing one that the other contexts cannot provide when you have structured data loss.

It is challenging to synthesize these results to propose a clear hierarchy of imputation strategies predicated on the correlation type being exploited. The findings demonstrate that the imputation context is not arbitrary, and is a critical modeling decision that encodes assumptions about the structure of the data; and second that when a subject has multi-joint data available, it is obviously better context for using the Across-Angle context because of its use of physically constrained. In contrast the Across-Player context is a useful alternative for single-angle analysis across a cohort maybe better than designated parameters in the context and for Across-Player context. Lastly, for univariate contexts - it is limited to that instant when there is no other correlated data, with the explicit understanding of its vulnerability to structured data loss.

\begin{figure}[H]
        \subfloat[Data imputation for single player\label{fig:random_points_errbar}]{
            \includegraphics[width=0.55\linewidth]{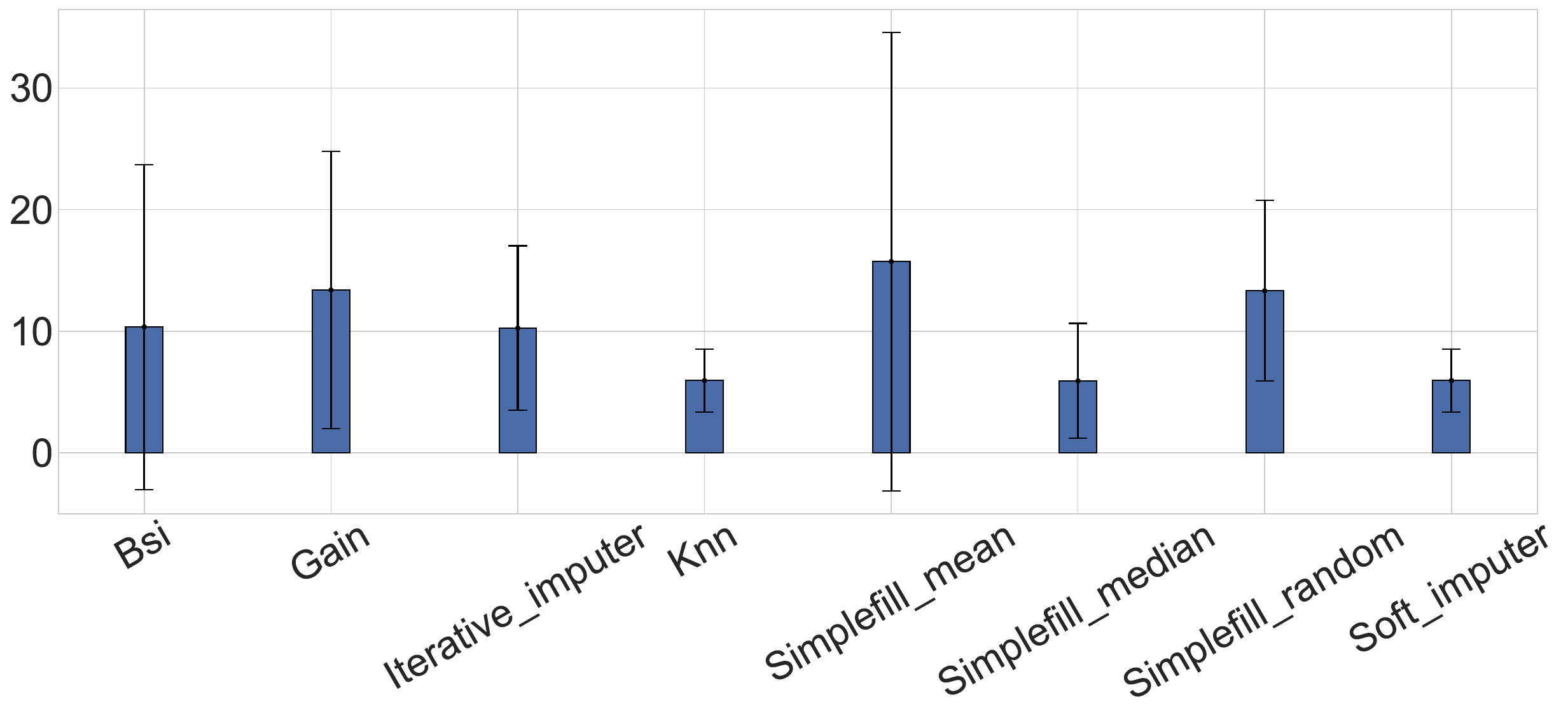}}
~
		\subfloat[Data imputation for multi-players of the same angle\label{fig:transition_points_errbar}]{
            \includegraphics[width=0.55\linewidth]{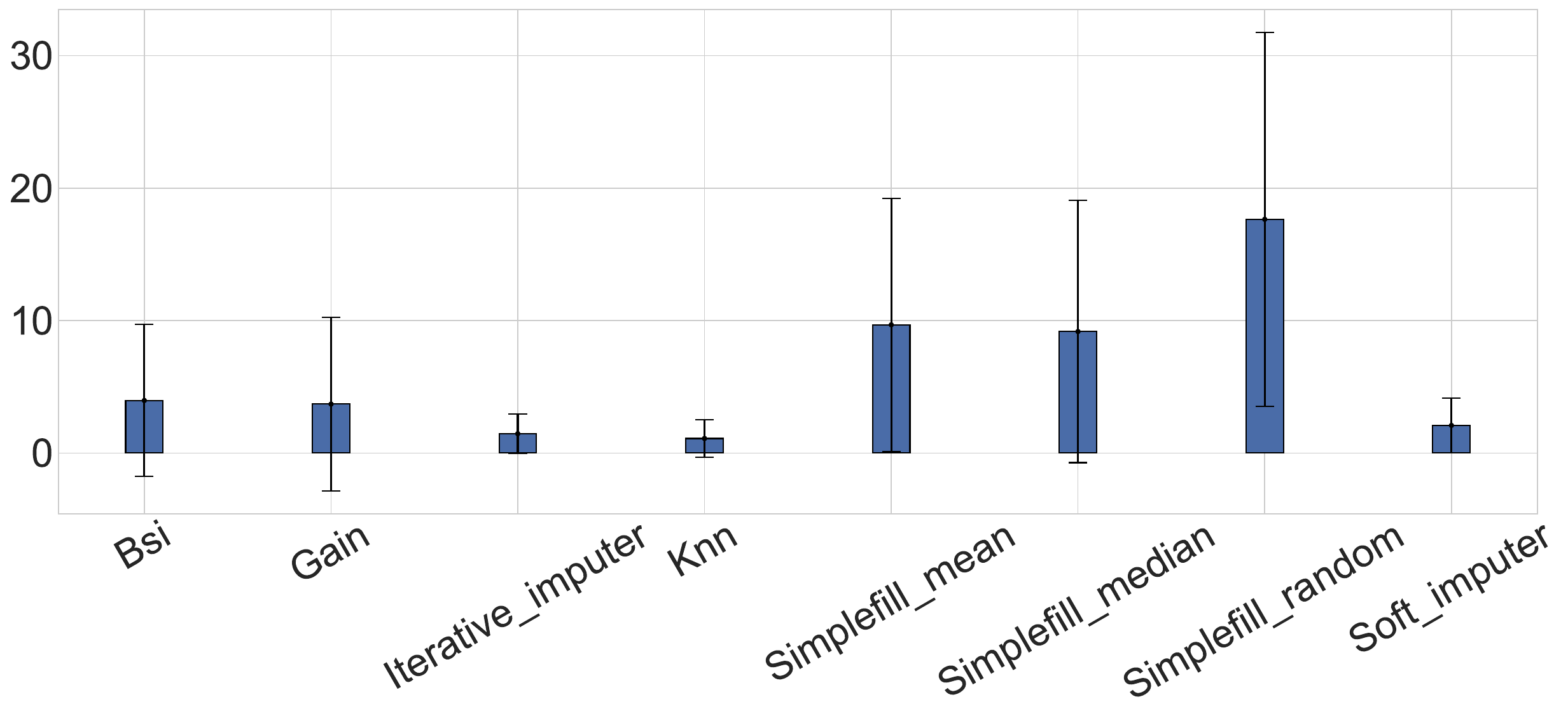}}

        \centering
        \subfloat[Data imputation for multi-angles of the same player\label{fig:interval_errbar}]{
            \includegraphics[width=0.55\linewidth]{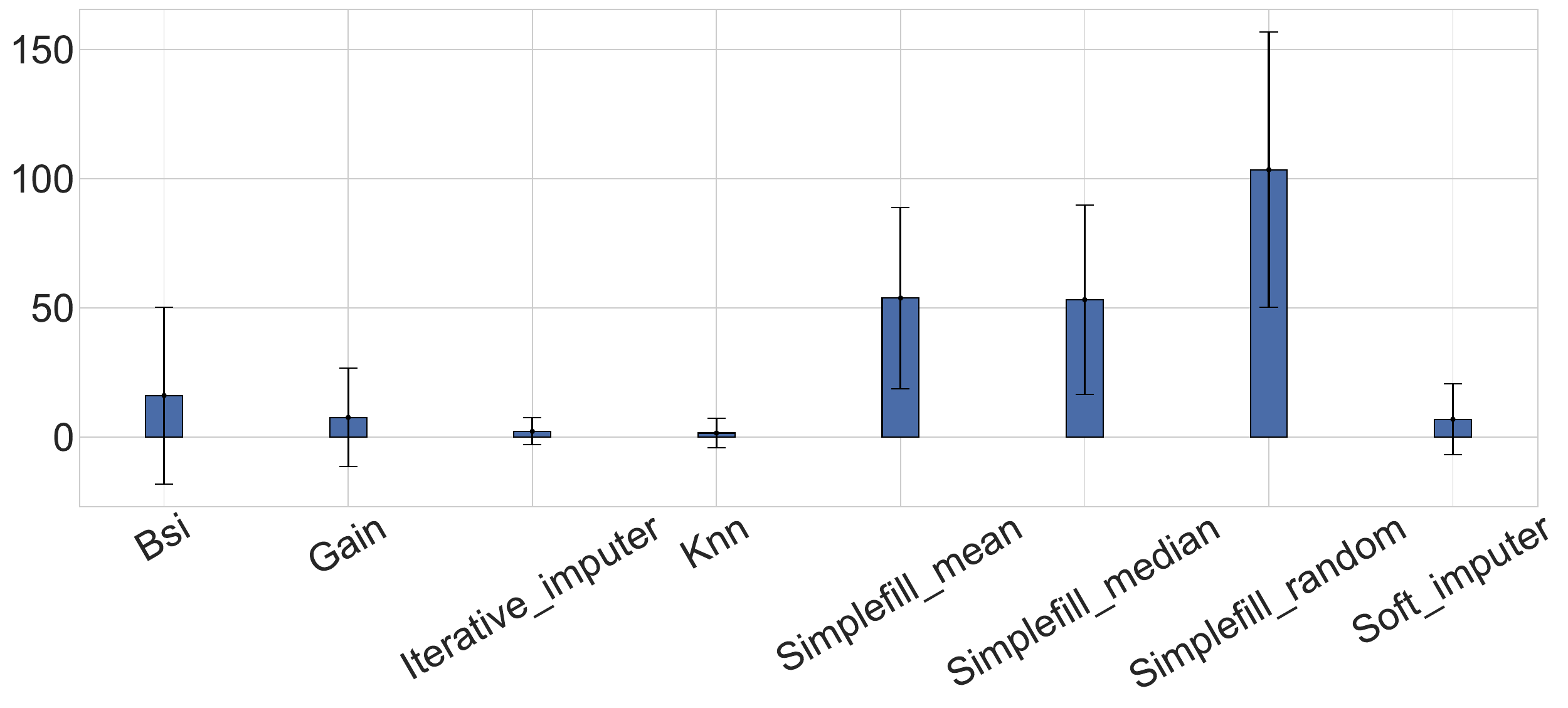}}
    \caption{MAE Error bar for data imputation of Transition missing points.
    % of a single data series. 
    \label{fig:Multiangle_errbar} }
\end{figure}

A direct comparison of the error bar plots for the transition point missingness mechanism, shows a significant difference in imputation performance across the three contexts and underlines the importance of the correlation of the data itself, as shown in Fig. \ref{fig:Multiangle_errbar}. In the univariate and multivariate multi-player contexts, advanced models (e.g., Iterative Imputer and KNN) achieve much better performance than basic statistical models to achieve a MAE of around 1-5. Due to large variance with Simplefill\_mean and Simplefill\_random, the error bars extend to MAE of over 30 in some cases. Thus, using  these statistical methods is risky for dynamically important data. However, performance characteristics shift dramatically when moving to the multi-angle context. In the multi-angle context, advanced models that can leverage biomechanically coupled kinematic variables, achieve an MAE around zero with very little variance (indicating perfect reconstruction), while simple statistical-based methods fail catastrophically, with Simplefill\_random demonstrating MAE over 100. This makes sense because they these methods perform a physically incongruous operation of averaging across distinct and dynamically different angle (e.g., shoulder and wrist), which results in a physically impossible data ranges and very large errors. This clearly demonstrates that just providing more data within a particular context and expecting to deal with it is not enough; we also need to consider the imputation approach chosen with data by which we can raise accuracy and trustworthiness in our results.

% ------------------------ Error bar

\section{Conclusions\label{sec:conclusions}} 
This work has provided a broad introduction to data imputation with missing data within the area of motion capture through IMU sensors, an area where data quality is important. We were able to perform the first exhaustive comparison of statistical methods, machine learning methods, and deep learning methods across three new imputation contexts, Univariate, Multivariate-Player, and Multivariate-Angle. The quantitative results indicate that the Multivariate-Angle context performed best across each of the contributors, and under this context the advanced IM methods such as the Iterative Imputer or KNN generated improvements in MAE reduction from 12.0±9.89 to 4.29±7.06 (64\% reduction) when learning from multi-player information. A key contribution of this paper is the introduction of the first publicly available benchmark dataset to help standardize the evaluation of imputation methods for this unique time-series data.  All our experimental findings are clear.  We show that multivariate frameworks, that exploit the correlations across players or kinematic variables, are far superior to univariate approaches, especially with unusual patterns of missingness. The more sophisticated models such as GAIN and the Iterative Imputer produced the highest quality data imputation - these are the best models to be able to capture the complex aspect of human motion.  The scale of the dataset (53 subjects, 100 time points) may limit the generalisability of our findings to sequences longer or with more diverse movements. The work we have undertaken has created a good baseline, and defined the most apparent next steps to support future researchers. Based on our findings, we suggest the following; (1) use Multivariate-Angle context when complete kinematic data is accessible, (2) use either Iterative Imputer or GAIN to manage complex patterns of missingness, and (3) avoidance of simple statistical methods for dynamic transition points as they can double or even triple the errors of the data estimates in the imputed data.  Future researchers should focus on transparent validation studies when working with real observations that will have naturally occurring gaps.

\section*{Funding Support}
This work was partially supported by a research grant from the Omani Ministry of Higher Education, Research, and Innovation under the project number BFP/RGP/ICT/23/382.

\bibliographystyle{plain} 
\bibliography{refs}

\begin{thebibliography}{10}

\bibitem{adel2022survey}
Basant Adel, Asmaa Badran, Nada~E Elshami, Ahmad Salah, Ahmed Fathalla, and Mahmoud Bekhit.
\newblock A survey on deep learning architectures in human activities recognition application in sports science, healthcare, and security.
\newblock In {\em The International Conference on Innovations in Computing Research}, pages 121--134. Springer, 2022.

\bibitem{adhikari2022comprehensive}
Deepak Adhikari, Wei Jiang, Jinyu Zhan, Zhiyuan He, Danda~B Rawat, Uwe Aickelin, and Hadi~A Khorshidi.
\newblock A comprehensive survey on imputation of missing data in internet of things.
\newblock {\em ACM Computing Surveys}, 55(7):1--38, 2022.

\bibitem{ae2018missing}
Jung Ae~Lee and Jeff Gill.
\newblock Missing value imputation for physical activity data measured by accelerometer.
\newblock {\em Statistical methods in medical research}, 27(2):490--506, 2018.

\bibitem{ahmad2013reviews}
Norhafizan Ahmad, Raja Ariffin~Raja Ghazilla, Nazirah~M Khairi, and Vijayabaskar Kasi.
\newblock Reviews on various inertial measurement unit (imu) sensor applications.
\newblock {\em International Journal of Signal Processing Systems}, 1(2):256--262, 2013.

\bibitem{aktakka2017chip}
Ethem~Erkan Aktakka, Jong-Kwan Woo, and Khalil Najafi.
\newblock On-chip characterization of scale-factor of a mems gyroscope via a micro calibration platform.
\newblock In {\em 2017 IEEE International Symposium on Inertial Sensors and Systems (INERTIAL)}, pages 1--4. IEEE, 2017.

\bibitem{arlotti2022benefits}
Jacob~S Arlotti, William~O Carroll, Youness Afifi, Purva Talegaonkar, Luciano Albuquerque, John~E Ball, Harish Chander, Adam Petway, et~al.
\newblock Benefits of imu-based wearables in sports medicine: Narrative review.
\newblock {\em International Journal of Kinesiology and Sports Science}, 10(1):36--43, 2022.

\bibitem{barker2011single}
Jamie Barker, Paul McCarthy, Marc Jones, and Aidan Moran.
\newblock {\em Single-case research methods in sport and exercise psychology}.
\newblock Routledge, 2011.

\bibitem{benson2021evaluating}
Lauren~C Benson, Carlyn Stilling, Oluwatoyosi~BA Owoeye, and Carolyn~A Emery.
\newblock Evaluating methods for imputing missing data from longitudinal monitoring of athlete workload.
\newblock {\em Journal of Sports Science \& Medicine}, 20(2):188, 2021.

\bibitem{cai2013accelerometer}
Qingzhong Cai, Ningfang Song, Gongliu Yang, and Yiliang Liu.
\newblock Accelerometer calibration with nonlinear scale factor based on multi-position observation.
\newblock {\em Measurement Science and Technology}, 24(10):105002, 2013.

\bibitem{clemente2021validity}
Filipe~Manuel Clemente, Zeki Akyildiz, Jos{\'e} Pino-Ortega, and Markel Rico-Gonz{\'a}lez.
\newblock Validity and reliability of the inertial measurement unit for barbell velocity assessments: A systematic review.
\newblock {\em Sensors}, 21(7):2511, 2021.

\bibitem{de2013multiple}
Moniek~CM de~Goeij, Merel van Diepen, Kitty~J Jager, Giovanni Tripepi, Carmine Zoccali, and Friedo~W Dekker.
\newblock Multiple imputation: dealing with missing data.
\newblock {\em Nephrology Dialysis Transplantation}, 28(10):2415--2420, 2013.

\bibitem{donders2006gentle}
A~Rogier~T Donders, Geert~JMG Van Der~Heijden, Theo Stijnen, and Karel~GM Moons.
\newblock A gentle introduction to imputation of missing values.
\newblock {\em Journal of clinical epidemiology}, 59(10):1087--1091, 2006.

\bibitem{Ette2006245}
Ene~I. Ette, Hui-May Chu, and Alaa Ahmad.
\newblock {\em Data Imputation}.
\newblock 2006.
\newblock Cited by: 2.

\bibitem{fathalla2023real}
Ahmed Fathalla, Ahmad Salah, Mahmoud Bekhit, Esraa Eldesouky, Ahmed Talha, Abdalla Zenhom, Ahmed Ali, et~al.
\newblock Real-time and automatic system for performance evaluation of karate skills using motion capture sensors and continuous wavelet transform.
\newblock {\em International Journal of Intelligent Systems}, 2023, 2023.

\bibitem{guignard2021validity}
Brice Guignard, Omar Ayad, H{\'e}lo{\"\i}se Baillet, Florian Mell, David Simba{\~n}a~Escobar, J{\'e}r{\'e}mie Boulanger, and Ludovic Seifert.
\newblock Validity, reliability and accuracy of inertial measurement units (imus) to measure angles: application in swimming.
\newblock {\em Sports Biomechanics}, pages 1--33, 2021.

\bibitem{hammerling2014completing}
Dorit Hammerling, Matthew Cefalu, Jessi Cisewski, Francesca Dominici, Giovanni Parmigiani, Charles Paulson, and Richard~L Smith.
\newblock Completing the results of the 2013 boston marathon.
\newblock {\em PLoS One}, 9(4):e93800, 2014.

\bibitem{hasan2021missing}
Md~Kamrul Hasan, Md~Ashraful Alam, Shidhartho Roy, Aishwariya Dutta, Md~Tasnim Jawad, and Sunanda Das.
\newblock Missing value imputation affects the performance of machine learning: A review and analysis of the literature (2010--2021).
\newblock {\em Informatics in Medicine Unlocked}, 27:100799, 2021.

\bibitem{jalata2021movement}
Ibsa~K Jalata, Thanh-Dat Truong, Jessica~L Allen, Han-Seok Seo, and Khoa Luu.
\newblock Movement analysis for neurological and musculoskeletal disorders using graph convolutional neural network.
\newblock {\em Future Internet}, 13(8):194, 2021.

\bibitem{jang2020deep}
Jong-Hwan Jang, Junggu Choi, Hyun~Woong Roh, Sang~Joon Son, Chang~Hyung Hong, Eun~Young Kim, Tae~Young Kim, Dukyong Yoon, et~al.
\newblock Deep learning approach for imputation of missing values in actigraphy data: Algorithm development study.
\newblock {\em JMIR mHealth and uHealth}, 8(7):e16113, 2020.

\bibitem{jekauc2012missing}
Darko Jekauc, Manuel V{\"o}lkle, Lena L{\"a}mmle, and Alexander Woll.
\newblock Missing values in sport scientific studies: A practical guide to multiple imputation with spss.
\newblock {\em Sportwissenschaft}, 42:126--136, 2012.

\bibitem{kang2014assessing}
Jian Kang, Yan Yuan, and Carolyn Emery.
\newblock Assessing remedies for missing weekly individual exposure in sport injury studies.
\newblock {\em Injury prevention}, 20(3):177--182, 2014.

\bibitem{karuzaki2021realistic}
Effie Karuzaki, Nikolaos Partarakis, Nikolaos Patsiouras, Emmanouil Zidianakis, Antonios Katzourakis, Antreas Pattakos, Danae Kaplanidi, Evangelia Baka, Nedjma Cadi, Nadia Magnenat-Thalmann, et~al.
\newblock Realistic virtual humans for cultural heritage applications.
\newblock {\em Heritage}, 4(4):4148--4171, 2021.

\bibitem{kato2025relationship}
Tsuyoshi Kato, Ryota Kasugai, and Kensuke Sakai.
\newblock Relationship between the total quality recovery scale and race performance in competitive college swimmers over two seasons.
\newblock {\em Sports}, 13(5):139, 2025.

\bibitem{kozlov2014imu}
Alexander Kozlov, Igor Sazonov, and Nina Vavilova.
\newblock Imu calibration on a low grade turntable: Embedded estimation of the instrument displacement from the axis of rotation.
\newblock In {\em 2014 International Symposium on Inertial Sensors and Systems (ISISS)}, pages 1--4. IEEE, 2014.

\bibitem{leurent2018sensitivity}
Baptiste Leurent, Manuel Gomes, Rita Faria, Stephen Morris, Richard Grieve, and James~R Carpenter.
\newblock Sensitivity analysis for not-at-random missing data in trial-based cost-effectiveness analysis: a tutorial.
\newblock {\em Pharmacoeconomics}, 36(8):889--901, 2018.

\bibitem{li2015multiple}
Peng Li, Elizabeth~A Stuart, and David~B Allison.
\newblock Multiple imputation: a flexible tool for handling missing data.
\newblock {\em Jama}, 314(18):1966--1967, 2015.

\bibitem{lin2020missing}
Wei-Chao Lin and Chih-Fong Tsai.
\newblock Missing value imputation: a review and analysis of the literature (2006--2017).
\newblock {\em Artificial Intelligence Review}, 53:1487--1509, 2020.

\bibitem{little2014joys}
Todd~D Little, Terrence~D Jorgensen, Kyle~M Lang, and E~Whitney~G Moore.
\newblock On the joys of missing data.
\newblock {\em Journal of pediatric psychology}, 39(2):151--162, 2014.

\bibitem{luo2022evaluating}
Yuan Luo.
\newblock Evaluating the state of the art in missing data imputation for clinical data.
\newblock {\em Briefings in Bioinformatics}, 23(1):bbab489, 2022.

\bibitem{mistler2017comparison}
Stephen~A Mistler and Craig~K Enders.
\newblock A comparison of joint model and fully conditional specification imputation for multilevel missing data.
\newblock {\em Journal of Educational and Behavioral Statistics}, 42(4):432--466, 2017.

\bibitem{muzellec2020missing}
Boris Muzellec, Julie Josse, Claire Boyer, and Marco Cuturi.
\newblock Missing data imputation using optimal transport.
\newblock In {\em International Conference on Machine Learning}, pages 7130--7140. PMLR, 2020.

\bibitem{nguyen2024filling}
Quang Nguyen and Gregory~J Matthews.
\newblock Filling the gaps: A multiple imputation approach to estimating aging curves in baseball.
\newblock {\em Journal of Sports Analytics}, 10(1):77--85, 2024.

\bibitem{olivares2009high}
Alberto Olivares, Gonzalo Olivares, JM~Gorriz, and J~Ramirez.
\newblock High-efficiency low-cost accelerometer-aided gyroscope calibration.
\newblock In {\em 2009 International Conference on Test and Measurement}, volume~1, pages 354--360. IEEE, 2009.

\bibitem{ramli2013roles}
MN~Norazian Ramli, Ahmad~Shukri Yahaya, NA~Ramli, Noor Faizah Fitri~Md Yusof, and MMA Abdullah.
\newblock Roles of imputation methods for filling the missing values: A review.
\newblock {\em Advances in Environmental Biology}, 7(12 S2):3861--3870, 2013.

\bibitem{rubin2004multiple}
Donald~B Rubin.
\newblock {\em Multiple imputation for nonresponse in surveys}, volume~81.
\newblock John Wiley \& Sons, 2004.

\bibitem{ser2016performance}
Gazel Ser, Siddik Keskin, and MCAN Yilmaz.
\newblock The performance of multiple imputations for different number of imputations.
\newblock {\em Sains Malaysiana}, 45(11):1755--1761, 2016.

\bibitem{singh2022efficient}
GN~Singh and Ashok~K Jaiswal.
\newblock Efficient imputation methods to handle missing data in sample surveys.
\newblock {\em Journal of Statistical Theory and Practice}, 16(3):40, 2022.

\bibitem{singh2024integrated}
M~Sheetal Singh, Khelchandra Thongam, Prakash Choudhary, and PK~Bhagat.
\newblock An integrated machine learning approach for congestive heart failure prediction.
\newblock {\em Diagnostics}, 14(7):736, 2024.

\bibitem{sohrabi2017accuracy}
Hamid Sohrabi and Saeed Ebadollahi.
\newblock Accuracy enhancement of mems accelerometer by determining its nonlinear coefficients using centrifuge test.
\newblock {\em Measurement}, 112:29--37, 2017.

\bibitem{solaro2018simulation}
N~Solaro, A~Barbiero, G~Manzi, and PA~Ferrari.
\newblock A simulation comparison of imputation methods for quantitative data in the presence of multiple data patterns.
\newblock {\em Journal of Statistical Computation and Simulation}, 88(18):3588--3619, 2018.

\bibitem{sterne2009multiple}
Jonathan~AC Sterne, Ian~R White, John~B Carlin, Michael Spratt, Patrick Royston, Michael~G Kenward, Angela~M Wood, and James~R Carpenter.
\newblock Multiple imputation for missing data in epidemiological and clinical research: potential and pitfalls.
\newblock {\em Bmj}, 338, 2009.

\bibitem{takahashi2017multiple}
Masayoshi Takahashi.
\newblock Multiple ratio imputation by the emb algorithm: Theory and simulation.
\newblock {\em Journal of Modern Applied Statistical Methods}, 16(1):34, 2017.

\bibitem{van2018accuracy}
Eline Van~der Kruk and Marco~M Reijne.
\newblock Accuracy of human motion capture systems for sport applications; state-of-the-art review.
\newblock {\em European journal of sport science}, 18(6):806--819, 2018.

\bibitem{wang2020implementing}
Chinchin Wang, Tyrel Stokes, Russell Steele, Niels Wedderkopp, and Ian Shrier.
\newblock Implementing multiple imputation for missing data in longitudinal studies when models are not feasible: a tutorial on the random hot deck approach.
\newblock {\em arXiv preprint arXiv:2004.06630}, 2020.

\bibitem{wang2022implementing}
Chinchin Wang, Tyrel Stokes, Russell~J Steele, Niels Wedderkopp, and Ian Shrier.
\newblock Implementing multiple imputation for missing data in longitudinal studies when models are not feasible: An example using the random hot deck approach.
\newblock {\em Clinical Epidemiology}, pages 1387--1403, 2022.

\bibitem{wang2020two}
Lei Wang, Yun Sun, Qingguo Li, Tao Liu, and Jingang Yi.
\newblock Two shank-mounted imus-based gait analysis and classification for neurological disease patients.
\newblock {\em IEEE Robotics and Automation Letters}, 5(2):1970--1976, 2020.

\bibitem{white2010avoiding}
Ian~R White, Rhian Daniel, and Patrick Royston.
\newblock Avoiding bias due to perfect prediction in multiple imputation of incomplete categorical variables.
\newblock {\em Computational statistics \& data analysis}, 54(10):2267--2275, 2010.

\bibitem{wood2005comparison}
Angela~M Wood, Ian~R White, Melvyn Hillsdon, and James Carpenter.
\newblock Comparison of imputation and modelling methods in the analysis of a physical activity trial with missing outcomes.
\newblock {\em International journal of epidemiology}, 34(1):89--99, 2005.

\bibitem{woods2024best}
Adrienne~D Woods, Daria Gerasimova, Ben Van~Dusen, Jayson Nissen, Sierra Bainter, Alex Uzdavines, Pamela~E Davis-Kean, Max Halvorson, Kevin~M King, Jessica~AR Logan, et~al.
\newblock Best practices for addressing missing data through multiple imputation.
\newblock {\em Infant and Child Development}, 33(1):e2407, 2024.

\bibitem{yahyamotion}
Muhammad Yahya.
\newblock Motion capture sensing techniques used in human upper limb motion: a review muhammad yahya, jawad ali shah, kushsairy abdul kadir, zulkhairi m. yusof, sheroz khan, arif warsi.

\bibitem{Yoon2018GAIN}
Jinsung Yoon, James Jordon, and Mihaela Schaar.
\newblock Gain: Missing data imputation using generative adversarial nets.
\newblock In {\em International conference on machine learning}, pages 5689--5698. PMLR, 2018.

\bibitem{zhang2004nonparametric}
Li-Chun Zhang.
\newblock Nonparametric markov chain bootstrap for multiple imputation.
\newblock {\em Computational statistics \& data analysis}, 45(2):343--353, 2004.

\bibitem{zhang2021low}
Xin Zhang, Changle Zhou, Fei Chao, Chih-Min Lin, Longzhi Yang, Changjing Shang, and Qiang Shen.
\newblock Low-cost inertial measurement unit calibration with nonlinear scale factors.
\newblock {\em IEEE Transactions on Industrial Informatics}, 18(2):1028--1038, 2021.

\bibitem{zhu2004real}
Rong Zhu and Zhaoying Zhou.
\newblock A real-time articulated human motion tracking using tri-axis inertial/magnetic sensors package.
\newblock {\em IEEE Transactions on Neural systems and rehabilitation engineering}, 12(2):295--302, 2004.

\end{thebibliography}
\end{document}